\documentclass{article}
\pdfoutput=1

\usepackage[accepted]{icml2020}

\usepackage{amsmath,amsfonts}
\usepackage{amssymb,amsopn}
\usepackage{bm} % bold symbol

\usepackage{graphicx}  %Required

\newcommand{\mbf}[1]{\mathbf{#1}} % vector, matrix
\newcommand{\bs}[1]{\boldsymbol{#1}} % boldsymbol

\newcommand{\rnn}{\operatorname{RNN}}
\newcommand{\softmax}{\operatorname{Softmax}}

\def\D{\mathcal{D}}
\def\loss{\mathcal{L}}

\newcommand{\argmin}{\operatornamewithlimits{argmin}}

\newlength{\widebarargwidth}
\newlength{\widebarargheight}
\newlength{\widebarargdepth}

\newcommand{\eat}[1]{}

\newcommand{\bx}{\mathbf{x}}

\newcommand{\bz}{\mathbf{z}}

\newcommand{\by}{\mathbf{y}}

\newcommand{\bh}{\mathbf{h}}
\newcommand{\bo}{\mathbf{o}}

\newcommand{\bW}{\mathbf{W}}
\newcommand{\bb}{\mathbf{b}}
\newcommand{\bI}{\mathbf{I}}

\newcommand{\real}{\mathbb{R}}
\newcommand{\bof}{\mathbf{f}}

\newcommand{\bv}{\mathbf{v}} 
\newcommand{\be}{\mathbf{e}} 
\newcommand{\bg}{\mathbf{g}} 
\newcommand{\bw}{\mathbf{w}} 

\newcommand{\bu}{\mathbf{u}}
\newcommand{\balpha}{{\bs\alpha}} 
 
\newcommand{\bgamma}{{\bs\gamma}} 

\newcommand{\bphi}{{\bs\phi}}
\newcommand{\bTheta}{{\bs\Theta}}
\newcommand{\btheta}{{\bs\theta}}

\usepackage[utf8]{inputenc} % allow utf-8 input
\usepackage[T1]{fontenc}    % use 8-bit T1 fonts
\usepackage{hyperref}       % hyperlinks
\usepackage{url}            % simple URL typesetting
\usepackage{booktabs}       % professional-quality tables
\usepackage{amsfonts}       % blackboard math symbols
\usepackage{nicefrac}       % compact symbols for 1/2, etc.
\usepackage{microtype}
\usepackage{graphicx}
\usepackage{subfigure}
\usepackage{booktabs} % for professional tables

\usepackage{multirow}
\usepackage{wrapfig}
\usepackage[skins]{tcolorbox}
\usepackage{makecell}
\usepackage{setspace}

\usepackage{mathtools}
\icmltitlerunning{Cost-effective Interactive Attention Learning with Neural Attention Process}

\begin{document}

\twocolumn[
\icmltitle{Cost-effective Interactive Attention Learning with Neural Attention Process}




\begin{icmlauthorlist}
\icmlauthor{Jay Heo}{kaist}
\icmlauthor{Junhyeon Park}{kaist}
\icmlauthor{Hyewon Jeong}{kaist}
\icmlauthor{Kwang joon Kim}{yonsei}
\icmlauthor{Juho Lee}{aitrics}
\icmlauthor{Eunho Yang}{kaist,aitrics}
\icmlauthor{Sung Ju Hwang}{kaist,aitrics}
\end{icmlauthorlist}

\icmlaffiliation{kaist}{Korea Advanced Institute of Science and Technology (KAIST), Daejeon, South Korea}
\icmlaffiliation{yonsei}{Yonsei University College of Medicine, Seoul, South Korea}
\icmlaffiliation{aitrics}{AITRICS, Seoul, South Korea}

\icmlcorrespondingauthor{Jay Heo}{jayheo@kaist.ac.kr}
% \icmlcorrespondingauthor{Junhyeon Park}{pjh2941@kaist.ac.kr}
% \icmlcorrespondingauthor{Hyewon Jeong}{jhw162@kaist.ac.kr}
% \icmlcorrespondingauthor{Kwang joon Kim}{preppie@yuhs.ac} 
% \icmlcorrespondingauthor{Juho Lee}{juho@aitrics.com}
% \icmlcorrespondingauthor{Eunho Yang}{eunhoy@kaist.ac.kr}
\icmlcorrespondingauthor{Sung Ju Hwang}{sjhwang82@kaist.ac.kr}

\icmlkeywords{Machine Learning, ICML}

\vskip 0.3in
]



\printAffiliationsAndNotice{}  % leave blank if no need to mention equal contribution

\begin{abstract}
We propose a novel interactive learning framework which we refer to as \emph{Interactive Attention Learning (IAL)}, in which the human supervisors interactively manipulate the allocated attentions, to correct the model's behavior by updating the attention-generating network. However, such a model is prone to overfitting due to scarcity of human annotations, and requires costly retraining. Moreover, it is almost infeasible for the human annotators to examine attentions on tons of instances and features. We tackle these challenges by proposing a sample-efficient attention mechanism and a cost-effective reranking algorithm for instances and features. First, we propose \emph{Neural Attention Process (NAP)}, which is an attention generator that can update its behavior by incorporating new attention-level supervisions without any retraining. Secondly, we propose an algorithm which prioritizes the instances and the features by their negative impacts, such that the model can yield large improvements with minimal human feedback. We validate IAL on various time-series datasets from multiple domains (healthcare, real-estate, and computer vision) on which it significantly outperforms baselines with conventional attention mechanisms, or without cost-effective reranking, with substantially less retraining and human-model interaction cost. 
\end{abstract}

\vspace{-0.1in}
\section{Introduction}
\vspace{-0.05in}
Deep neural networks are arguably the most prevalent tools for predictive modeling tasks nowadays, thanks to their ability to learn complex functions with multiple layers of non-linear transformations.
However, the complex nature of the model, at the same time, makes it difficult to interpret what they have learned, which has led to the recent surge of interest in interpretable models that are capable of providing interpretations of the model and the prediction in human-understandable forms~\citep{gilpin2018explaining}. 

Although recent works propose diverse solutions to interpretability~\citep{retain, ahmad2018interpretable, lage2018human}, including attention mechanisms, activation visualization, and optimization for human-interpretability under human-in-the-loop, we face yet another challenge: not all machine-generated interpretations are \textit{correct} or \textit{human understandable}. This is mainly due to two reasons: 1) correctness and reliability of a learning model heavily depends on the quantity and quality of the training data. 2) neural networks tend to learn \emph{non-robust} features that help with predictions but are not human-perceptible~\citep{ilyas2019}. Such unreliability of the interpretations is highly problematic for safety-critical applications such as clinical risk predictions~\citep{ahmad2018interpretable, sankar2019sisc} or autonomous driving~\citep{chi2017deep}.

The main limitation of the existing models is that they mostly only consider passive roles for human supervisors, where they simply take the provided interpretations as is. Yet, a more effective way to use the interpretations is to use them as channels for human-model communications, such that the models learn by continuously \emph{interacting} with the human supervisors, where they iteratively correct the model-generated interpretations. From a cognitive science perspective, human learning is done by internal reflection (\emph{back-propagation}) and external explanation (\emph{human feedback}) during social interactions~\citep{neuroscience}. 

\begin{figure*}[t]
\begin{center}
\small
\begin{tabular}{c}
\hspace{0.25in}
\includegraphics[width=16cm, height=3.4cm]{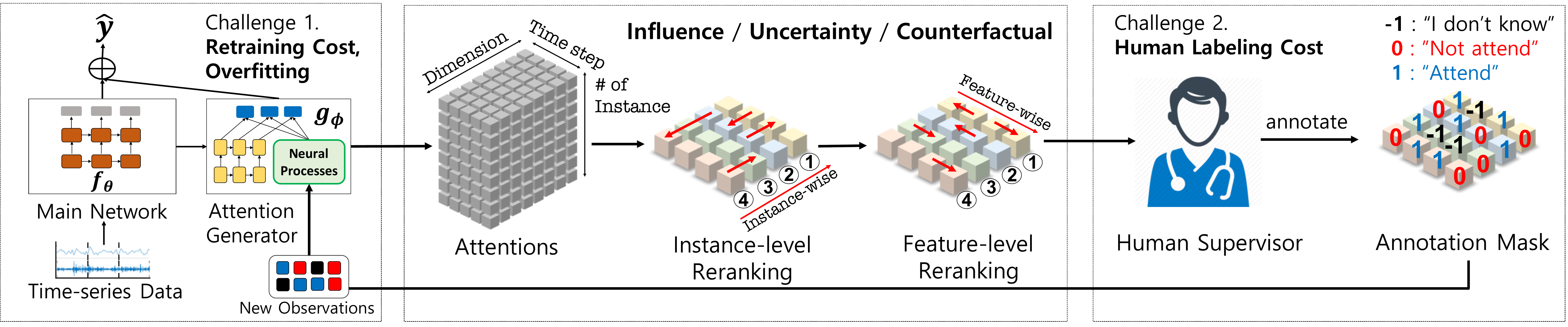}\\
\end{tabular}
\end{center}
\vspace{-0.1in}
\small
\begin{tabular}{c c c}
\hspace{0.32in}(A) Neural Attention Process & \hspace{0.60in} (B) Cost-Effective Re-ranking & \hspace{0.75in} (C) Human Annotation \\
\end{tabular}
\vspace{-0.14in}
\caption{\small \textbf{Our Interactive Attention Learning (IAL) framework.} IAL is an interactive learning framework which iteratively learns by interacting with the human supervisor, via the learned attentions. It allows efficient model update using (A) Neural Attention Process which does not require retraining, and cost-effective interaction via (B) Cost-effective reranking of the instances and features.
}
%In a typical interactive learning setting, there exists two challenges: 1) \textbf{Expensive retraining cost and overfitting} - retraining is expensive with neural networks with millions of parameters, and training with scarce data could result in overfitting, 2)\textbf{Expensive human labeling cost} - Requesting and obtaining human feedback is costly with a large number of datasets. A good interactive learning framework needs to cost-effectively minimize labor cost and, at the same time, efficiently relearn with scarce data without expensive retraining cost.}
\label{ial_framework}
\vspace{-0.16in}
\end{figure*}
%To address such problems, interactive learning framework could be effective means by embedding human knowledge into a learning model: human can interactively modify the training process of neural network itself such that the model learns what to learn or ignore for given input features. This interactive coordination between machine and human corresponds to the perspective of cognitive neuroscience, in that human learning and understanding are built upon two integral parts: \emph{interaction} and \emph{explanation}. Our brain's biological functions are constantly developed by internal reflection (\textit{Back-propagation}) and external explanation (\textit{Human feedback}) during social interactions~\citep{neuroscience}. 

Based on this motivation, we propose an interactive learning framework, where the model learns by iteratively interacting with the human supervisors who manipulate the model by adjusting the provided interpretations, which is depicted in Figure~\ref{ial_framework}. The specific interpretation mechanism we consider in this work is the \emph{attention} mechanism~\cite{bahdanau2014neural}. While active learning asks for supervision at the instance level, in our interactive learning model, it asks for supervision at the attention level. However, this leads to multiple challenges regarding efficiency, which hinders their applications to practical scenarios:
\vspace{-0.15in}
\begin{itemize}
    \item \textbf{Model retraining cost and overfitting:} To reflect human feedback, the model needs to be retrained, which is costly. Moreover, retraining the model with scarce human feedback may result in the model overfitting.
    % \vspace{-0.1in}
    \item \textbf{Expensive human supervision cost:} Obtaining human feedback on datasets with large numbers of training instances and features is extremely costly. Further, obtaining feedback on already correct interpretations is wasteful.
    \vspace{-0.1in}
\end{itemize}

To tackle these practical challenges, we propose a novel interactive learning framework, which we refer to as \emph{Interactive Attention Learning (IAL)}, that allows both efficient model retraining and sample-efficient learning that minimizes human supervision cost. IAL consists of two main components: 1) \textbf{Neural Attention Processes (NAP)} and 2) \textbf{Cost-Effective instance and feature Reranking (CER)}. Basically, our model minimizes retraining cost via NAP which allows the model to correct its attention-generating behaviour in a sample-efficient manner by incorporating new labeled instances without retraining. NAP also prevents overfitting, which is inevitable with scarce human feedbacks when using a conventional attention mechanism. Secondly, to address the expensive human labeling cost, \textbf{CER} reranks the instances, features, and timesteps (for time-series data) by their negative impacts. This enables the model to minimize human interaction cost, such that the human supervisors only correct the interpretations that are likely to be incorrect and influential to the prediction. The importance of each sample and feature is measured either by the uncertainty, influence function~\citep{cook1980characterizations}, or counterfactual estimation. 

We validate our IAL framework on a variety of real world tasks with time-series data, including cerebral infarction risk prediction from electronic health records (EHR), New York City real-estate price forecast, and squat-posture prediction task. The experimental results show that our model outperforms baseline interactive learning schemes with significant margins, with considerably smaller interaction cost in terms of both model retraining and human annotation cost.
%~\footnote{We will release the source codes and the datasets upon acceptance of our paper}
Our contributions are as follows:
\vspace{-0.1in}
\begin{itemize}
\item We propose a \textbf{novel interactive learning framework} which iteratively updates the model by interacting with the human supervisor via the generated attentions.

\item To minimize the retraining cost, we propose a \textbf{novel probabilistic attention mechanism} which sample-efficiently incorporates new attention-level supervisions on-the-fly without retraining and overfitting. 

%\item We propose cost-effective selection solutions that can rank the examples and the attended input variables, in order to maximize the effect of each annotation and thus to minimize the human-machine interaction cost.
\item To minimize human supervision cost, we propose an \textbf{efficient instance and feature reranking} algorithm, that prioritizes them based on their negative impacts on the prediction, measured either by uncertainty, influence function, or counterfactual estimation.   

%\item We propose a novel interactive attention learning framework with an efficient attention mechanism based on Neural Process, which enables to efficiently correct the model's understanding with scarce human feedback without retraining of the entire network. 
\item We validate our model on \textbf{five real-world datasets} with binary, multi-label classification, and regression tasks, and show that our model obtains significant improvements over baselines with substantially less retraining and human feedback cost.
\end{itemize}
\vspace{-0.2in}
%\vspace{-0.01in}
\section{Related work}
%\paragraph{Interactive Learning} 
\paragraph{Interpretable machine learning} 
The literature on interpretable machine learning is vast, but we only discuss a few. A popular approach to obtain interpretable model is to build a simple proxy model that mimics the (local) behaviours of a complex model, using either simplified linear models~\citep{Ribeiro2016} or decision trees~\citep{sato2001rule, salzberg1994c4}. Another approach, specific for neural networks, is analyzing their learned representations~\citep{sharif2014cnn, yosinski2014transferable} at each unit via visualization. ~\citet{bau2017network} further consider interpretability of representations in light of their correspondence to semantic concepts, and utilize it for controlling the behaviours of generative adversarial networks~\citep{bau2019visualizing}. In this work, we propose a novel interactive learning framework that leverages the model's interpretation to iteratively correct the model's behaviour, while minimizing the interaction cost.
%\vspace{-0.1in}
% \paragraph{Attention Mechanism}
% Attention mechanism~\cite{bahdanau2014neural} is an effective approach to adaptively select a subset of features (or inputs) in an input-dependent manner, such that the model dynamically focuses on more relevant features for prediction. This mechanism works by input-adaptively generating coefficients for the input features to locate more weights to the features that are more relevant for the prediction on the given input. Attention mechanisms have achieved success with various applications, including image translation~\citep{show_attend_tell}, machine translation~\citep{Bahdanau15}, memory-augmented networks~\citep{e2ememnet}, and visual question answering~\citep{das2017human}, and healthcare~\citep{retain}. \vspace{-0.2in} %In this work, we consider attention as a way to both understand what the model has learned and to efficiently correct the model's behaviour. %We also propose a novel data-efficient attention mechanism based on Neural Process, which generalizes well with scarce human labels and can incorporate new labeled instances without retraining.

\paragraph{Attention Mechanism}
Attention mechanism~\cite{bahdanau2014neural} is an effective approach to adaptively select a subset of features in an input-dependent manner, such that the model dynamically focuses on more relevant features for prediction. This mechanism works by input-adaptively generating coefficients for input features to allocate more weights to more relevant features for prediction. Attention mechanisms have achieved success with various applications, including image translation~\citep{show_attend_tell}, natural language understanding~\citep{Bahdanau15, vaswani2017attention}, and visual question answering~\citep{das2017human}. However, in the interactive learning setting, conventional attention mechanisms are either not trainable, or require retraining of the attention generator on the newly delivered attention-level annotations, which may lead to performance degeneration due to catastrophic forgetting. In this work, we incorporate benefits from the nonparametric and amortized inference of Neural Process (NPs)~\citep{Garnelo2018np} into an attention mechanism such that it generalizes well with scarce human labels in a semi-supervised manner and can incorporate new labeled instances without retraining via an approximation of stochastic process.

% \paragraph{Neural Processes} Neural Processes (NPs) is a neural network-based formulation that combine benefits of deep neural network and stochastic process, which learns an approximation of a stochastic process~\citep{Garnelo2018np}. NPs allow for global sampling via a latent variable \textbf{z} to produce different function samples and model the uncertainty for some given context data. \citep{conditional_np} introduced Conditional Neural Processes (CNPs) which differs from NPs, in that CNPs do not sample different functions for the same context points, since it doesn't generate a latent variable for global sampling. \citep{attentive_np} resolves the underfitting problem caused by the mean-aggregator, by utilizing attention mechanism. Our data-efficient attention mechanism based on Neural Processes benefits from the nonparametric and amortized inference of NPs, which allows for incorporating new labeled samples without retraining. 

\paragraph{Active learning} While there are vast literature on annotation methodology and active learning~\citep{tong2001active, sener2017active}, we here discuss a few relevant pre-existing works for learning from rationales, which is a popular annotation technique in natural language processing~\citep{zaidan2008modeling} and vision~\citep{donahue2011annotator}, where a human highlights the important region of input. However, while these works directly zero out or modify input features, the attention generator in IAL provides its interpretation in the form of the attention, and the human supervisor corrects them. Furthermore, in conventional active learning settings, annotators’ roles are relatively \textit{passive}, as they simply provide labels to each given instance such that they can't see the effect of one’s annotation. However, the annotators in IAL \textit{actively} interpret the generated attentions, \textit{directly} modify the learning manifold of the model by masking them, and can \textit{immediately} see the effect of the newly added annotation.

\newcommand{\ssc}[1]{{\scriptscriptstyle{#1}}}
\newcommand{\tth}[1]{^{\ssc{(#1)}}}
%\vspace{-0.06in}
\section{Interactive Attention Learning}
Suppose we have a pre-trained neural network $\mbf{F}_{\bTheta}$ with a parameter $\bTheta$ trained on a dataset $\mathcal{D}_\text{train} = \{(\bx_i\tth{1:T}, \by_i)\}_{i=1}^N$. $\bx_i\tth{1:T} = [\bx_i\tth{1}, \dots, \bx_i\tth{T}]$ is a time-series instance with $\bx_i\tth{t} \in \real^{D}$, and $\by_i \in \real^{L}$ is the corresponding label. We denote each labeled instance as $\bu_i = (\bx_i\tth{1:T}, \by_i)$. $\bTheta$ is trained to minimize the empirical risk, the expectation of individual loss $\loss(\bTheta, \bu_i)$ over all training instances; we use mean-squared error for regressions or the categorical cross-entropy for classification problems. We further assume that $\bTheta$ consists of two sub-parameters $(\btheta, \bphi)$, where $\btheta$ corresponds to the parameter of the main neural network $\bof_\btheta$ and $\bphi$ corresponds to the parameter of \emph{the attention-generating network} $\bg_\bphi$. $\bg_\bphi$ generates an attention $\balpha\tth{1:T}_i$ for $\bx_i\tth{1:T}$, where each $\balpha\tth{t}_i$ is separated into an attention for time-axis $\beta\tth{1:T}_i$ and an attention for feature-axis $\bgamma\tth{1:T}_i$ (see \eqref{eq:retain_attn} for detailed definition). The attentions are applied to the $D$ features along $T$ time-steps, and let the model focus on a specific features of the representations of inputs relevant to the prediction. Hence, the attention provides an interpretation of the model's decision.

Our goal in this paper is to correct the behaviour of the attention-generating network $\bg_{\bs\phi}$ with human supervision. This may be done by incrementally retraining $\bg_{\bs\phi}$ over multiple rounds, where for each round human supervisors inspect the attentions generated by $\bg_{\bs\phi}$ and update $\bs\phi$. We assume that a human supervisor provides an \emph{attention mask} $\mbf{m}\tth{1:T}_{i}$ for each sample $\bx\tth{1:T}_{i}$ as ground-truth label, after \emph{manually examining} the attention $\balpha\tth{1:T}_{i}$ produced by $\bg_\bphi$. An attention mask for a certain axis is defined to be a ternary value $\{-1, 0, 1\}$, where $-1$ indicates \textit{"I don't know"}, $0$ indicates \textit{"Not attend"}, and $1$ indicates \textit{"Attend"}. Note that a na\"ive retraining of $\bg_\bphi$ leads to the costly retraining of $\bof_\btheta$ via gradient back-propagation. Instead, we choose to fix $\btheta$ and update $\bs\phi$ only to minimize the cost of retraining. We refer to this general framework that learns by interacting with the human supervisor via learned attention, as \emph{Interactive Attention Learning framework (IAL)}.

% \begin{algorithm}[tb]
%     \small
%     \caption{Interactive Attention Learning Framework}
%     \label{alg:ial_algorithm}
%     \textbf{\bfseries Input:} $\D_\text{train} = \{\bx_i\tth{1:T}, \by_i\}_{i=1}^{N}$, $\bm\Theta = \{\bm\omega, \bm\phi\}$, $S$.\\
%     \textbf{\bfseries Output:} $\bm\Theta$.   
%     \begin{algorithmic}
%         \FOR{$s=1,...,S$}
%             \IF{$s=1$}
%                 \STATE Train the network weights $\bm\Theta\tth{1}$ by \\ $\minimize_{\bm\Theta\tth{1}} \loss(\bm\Theta\tth{1}; \D_\text{train}) + \Omega(\bs\Theta\tth{1})$
%             \ELSE
%                 \STATE $(\D_\text{selected}\tth{s} = \{\bx\tth{1:T}_k, \by_k\}_{k=1}^{K}, \bs\alpha)$ $\leftarrow \mathrm{CESR}(\bm\theta^{(t-1)})$ \\ $\quad\quad$ $\triangleright$ Cost-Effective Selection (CES)
%                 \STATE $\{\mbf{m}_k\}^{K}_{k=1} \leftarrow$ $\mathrm{Evaluate}(\D_\text{selected}^{(s)} ,\bs\alpha)$ \\
%                 $\quad\quad$ $\triangleright$ Evaluate \& get feedback for attention $\bs\alpha$
%                 \STATE $\bs\phi\tth{s} \leftarrow \mathrm{NAP}(\bx_s, \{\mbf{m}_k\}^{K}_{k=1}, \bm\phi)$\\ 
%                 $\quad\quad$ $\triangleright$ Learn human feedback with quick forward \\
%                 $\quad\quad$ pass using Neural Attention Process (NAP).
%             \ENDIF
%         \ENDFOR
%     \end{algorithmic}
% \end{algorithm}

\begin{algorithm}[t]
    \small
    \caption{Interactive Attention Learning Framework}
    \label{alg:ial_algorithm}
    \textbf{\bfseries Input:} $\D_\text{train} = \{\bx_i\tth{1:T}, \by_i\}_{i=1}^{N}$, $\bm\Theta = \{\bm\theta, \bm\phi\}$, rounds $S$.\\
    \textbf{\bfseries Output:} $\bm\Theta$.
    \begin{algorithmic}[1]
        \STATE Pretrain $\bTheta\tth{0} = \argmin_{\bTheta} \loss(\bTheta, \D_\text{train}) + \Omega(\bTheta)$.
          \FOR{$s=1,...,S$}
                \STATE $\D_\text{selection}\tth{s}, \{\balpha\tth{1:T}_k\}_{k=1}^K = \mathrm{CER}(\bTheta\tth{s-1})$. \\
                $\quad\quad\triangleright$ Cost-Effective Re-ranking (CER)
                \STATE $\{\mbf{m}\tth{1:T}_k\}^{K}_{k=1} = \mathrm{Evaluate}(\D_\text{selection}^{(s)} ,\{\bs\alpha\tth{1:T}_{k}\}_{k=1}^{K})$ \\
                $\quad\quad$ $\triangleright$ Get attention masks for $\bs\alpha$
                \STATE $\bphi\tth{s} = \mathrm{NAP}(\D_\text{selection}^{(s)}, \{\mbf{m}\tth{1:T}_k\}^{K}_{k=1}, \bm\phi\tth{s-1})$\\
                $\quad\quad$ $\triangleright$ Learn human feedback with quick forward \\
                $\quad\quad$ pass using Neural Attention Process (NAP).
                \IF{$s=1$}
                \STATE Retrain $\bTheta\tth{1} = \argmin_\bTheta \loss(\bTheta, \D_\text{train}) + \Omega(\bTheta)$ with an adapted network containing NAP.
                \ENDIF
            \ENDFOR
    \end{algorithmic}
\end{algorithm}

\begin{figure*}[t]
\begin{center}
\small
\begin{tabular}{c c c}
%\hspace{-0.21in}
\includegraphics[width=3.4cm, height=3.1cm]{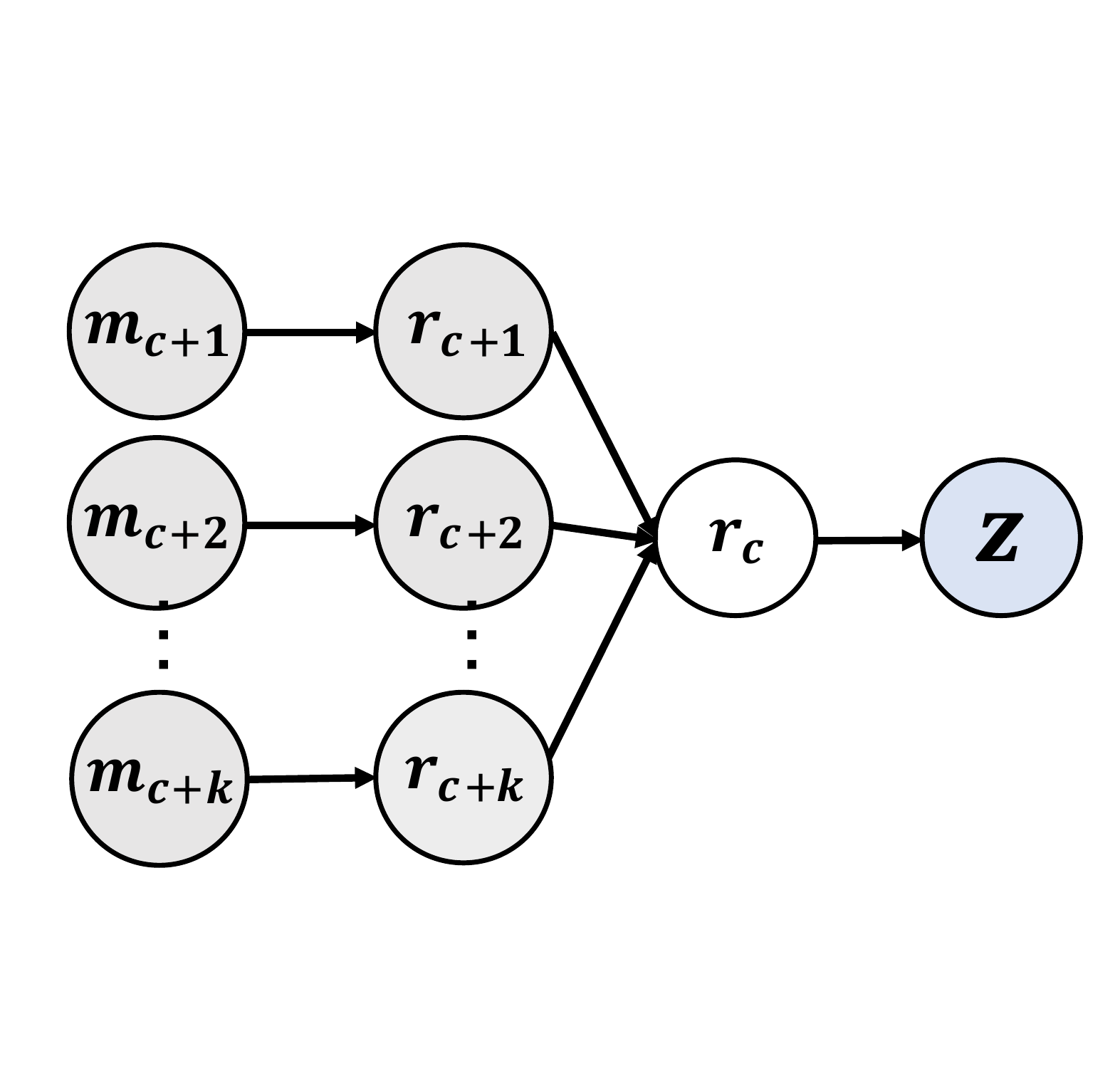} &
%\hspace{0.04in}
\includegraphics[width=3.9cm, height=3.2cm]{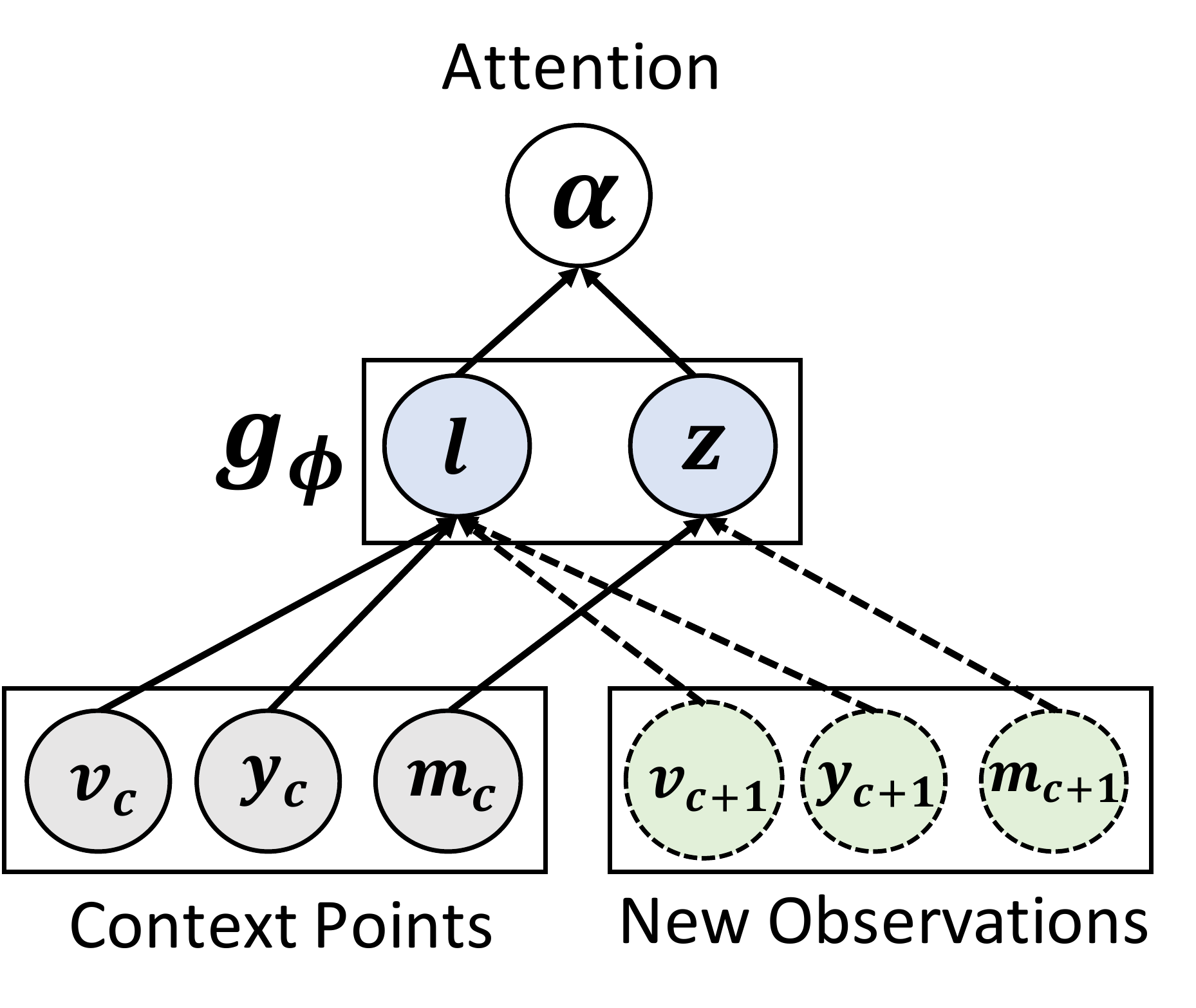} &
%\hspace{0.04in}
\includegraphics[width=5.3cm, height=3.2cm]{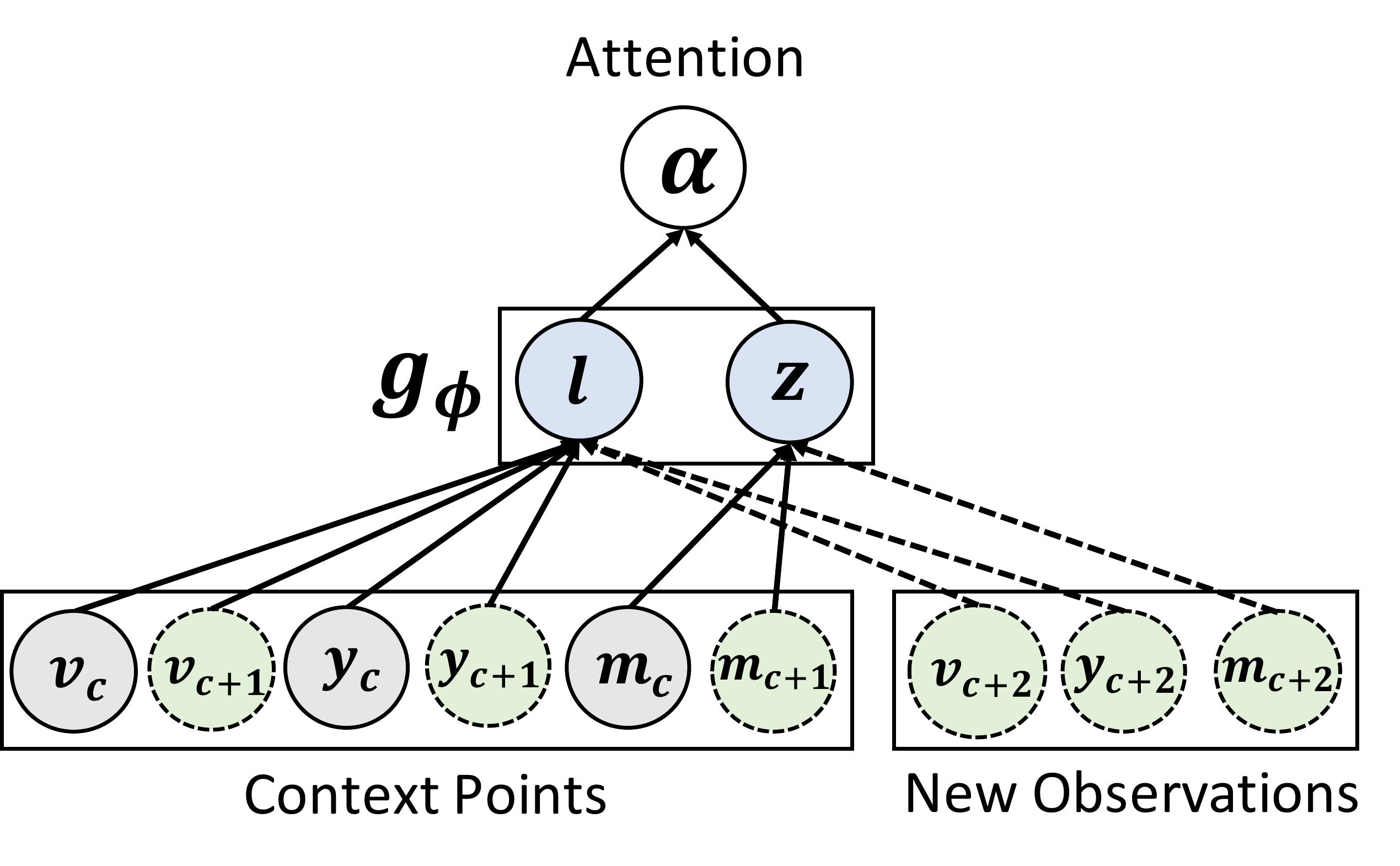}  \vspace{-0.06in}\\
\vspace{-0.02in}
\small (a) Neural Attention Process (NAP) & (b) First Round ($s$=1) & (c) Further Rounds ($s$=2,3,..) \\ 
%\textcolor{blue}{figure will be replaced} & & \\ 
\end{tabular}
\end{center}
\vspace{-0.16in}
\caption{\small (a): NAP naturally reflects the information from the annotation summarization $\bz$ via amortization. (b) For new observations (annotation mask $\mbf{m}_{c+1}$), NAP accepts them as input and generates the mean and variance parameter for $\bz$. (c) NAP doesn't require retraining for further new observations, in that NAP automatically adapt to them at the cost of a forward pass through a network $\mbf{g}_{\bs\phi}$.}
\label{fig:nap_concept}
%\vspace{-0.20in}
\end{figure*}
Yet, as discussed in the introduction, there are still remaining challenges that need to be tackled. First, the retraining of $\mbf{g}_{\bs\phi}$ will still incur a non-negligible cost and may also result in overfitting when human feedback is scarce. To tackle this, we propose a novel attention generator that can readily incorporate human annotations without retraining. Another challenge is reducing the human interaction cost. Ideally, a human annotator may have a look on the entire attentions generated by $\bg_\bphi$. This involves examining all instances $(\bu_i,\dots, \bu_N)$, and within each instance, all features over all time-steps $(\bu_{i,1}^{(1)}, \dots, \bu_{i,D}^{(T)})$. This is not feasible and wasteful since many attention values are already correct. To tackle this problem, we further propose a cost-effective reranking method which prioritizes the instances and features by their impacts on the model's prediction, to maximize performance gains with minimal human effort.

Algorithm~\ref{alg:ial_algorithm} describes the detailed algorithm for our IAL framework that leverages the proposed attention mechanism and re-ranking method. In the next two subsections, we describe the two components that minimize both the model retraining cost and human-model interaction cost.

\subsection{Neural Attention Process}
In this section, we describe \emph{Neural Attention Process (NAP)}, an novel attention generator based on NPs \citep{Garnelo2018np}. NAP can effectively update the model without retraining by amortization using sparse human annotations.

Before describing our approach, we briefly explain how attention is applied for time-series prediction, using RETAIN~\cite{retain} as our base model. 
Let $\bv\tth{1:T} = \bW_\text{emb} \bx\tth{1:T}$ be a linear embedding of an input. We restrict $\bv\tth{1:T}$ to have the same dimensionality $(D)$ as $\bx\tth{1:T}$,
so that we can directly compute the contribution of a certain feature to a prediction\footnote{Please refer to the supplementary material to see how to compute the contribution
of input features to predictions based on attentions and embedding $\bv\tth{1:T}$. For now, treat each dimension of $\bv\tth{1:T}$ to be directly linked
to the corresponding feature in $\bx\tth{1:T}$.}. The model computes attention coefficients for both \emph{time-steps} and \emph{input-features}
as,
% \vspace{-0.02in}
\begin{align}
\small
&\bo\tth{1:T} = \rnn_\beta(\bv\tth{1:T}),\\
&\bh\tth{1:T} = \rnn_\bgamma(\bv\tth{1:T}),\\
& e\tth{t} = \bw_\beta^\top \bo\tth{t} + b_\beta \text { for } t=1,\dots, T, \\
& \mathbf{q}\tth{t} = \bW_\bgamma \bh\tth{t} + \bb_\bgamma \text{ for } t=1,\dots, T, \\
&\beta\tth{1:T} = \softmax(e\tth{1}, \dots, e\tth{T}), \\
& \bgamma\tth{t} = \tanh(\mathbf{q}\tth{t}) \text{ for } t=1,\dots, T.
\label{eq:retain_attn}
\end{align}
\vspace{-0.02in}Here, $\beta\tth{1:T}$ are attention weights applied for time-steps and $\bgamma\tth{1:T}$ are attention weights for the input features. We may also consider the stochastic attention as in \citep{show_attend_tell}. Given $\bs\alpha\tth{1:T} = \{ \beta\tth{1:T}, \bgamma\tth{1:T}\}$, the model makes predictions as $\hat\by = \bh(\sum_{t=1}^T \beta\tth{t} \cdot ( \bgamma\tth{t} \odot \bv\tth{t} ))$ where $\odot$ is the element-wise multiplication and $\bh$ is an output layer.

Now we describe NAP, especially how it amortizes the procedure of updating the model given human annotations. Let $\{\mbf{m}_k\tth{1:T}\}_{k=1}^K$ be a set of attention masks given by human annotators for a subset $\D_\text{selection} = \{ (\bx_k\tth{1:T}, \by_k) \}_{k=1}^K \subseteq \D_\text{train}$ with $K \ll N$. Instead of exhaustively retraining $\bg_\bphi$, NAP learns to \emph{summarize} $\D_\text{selection}$ to a latent vector, and give the summarization as an additional input to the attention generating network. This approach, when trained properly, can automatically adapt to new annotations without having to retrain the parameters.
From below, we describe the components of NAP in more detail. 

\iffalse
Now we describe the actual algorithm for neural attention process. Let $\{\mbf{m}\tth{1:T}_k\}_{k=1}^K$ be a set of attention masks given by human annotaors. annotations represented as masks, given for the selected subsamples $\D_\text{selection} = \{ (\bx_i\tth{1:T}, \by_i) \}_{k=1}^K$ of the training data $\D_\text{train} = \{ (\bx_i\tth{1:T}, \by_i) \}_{i=1}^N$. Given that $N \gg K$, the core idea of \textbf{NAP} is that, instead of updating the parameter $\bs\phi$ using these small number of examples, we let network take the \emph{summarization} of the annotation set as an additional input. This approach, when trained properly, can automatically adapt without retraining when a new set of annotations are further given.
\fi

%The overall pipeline of neural attention process is depicted in Figure~\ref{fig:nap}.
\newcommand{\barr}{\bar{\mbf{r}}}
% \paragraph{Embedding the inputs} we first embed the input $\bx\tth{1:T}$ using LSTM~\citep{Hochreiter1997} into
% $\mbf{l}\tth{1:T} = [\mbf{g}\tth{1:T}, \mbf{h}\tth{1:T}]$. 

\paragraph{Embedding \& summarizing the annotations} 
We first feed the input embedding $\bv\tth{1:T}$ to LSTM~\citep{Hochreiter1997} ($\mathrm{RNN}_\beta, \mathrm{RNN}_\bgamma$) to generate time-series representation $\mbf{l}\tth{1:T} = [\mbf{\bo}\tth{1:T}, \mbf{h}\tth{1:T}]$. Given attention masks $\{\mbf{m}_k\tth{1:T}\}_{k=1}^K$, we build an intermediate representation $\{\mbf{r}\tth{1:T}_k\}_{k=1}^K$ via another LSTM. Then, for each time step, we build a summarized representation $\barr\tth{t}$ by a permutation-invariant operation (for instance, average),
\vspace{-0.01in}
\begin{align}
\small
\barr\tth{t} = \mbf{r}\tth{t}_1 \oplus \dots \oplus \mbf{r}\tth{t}_K.
\end{align}
Having $\barr\tth{1:T}$, we define a distribution for the summary variable $\bz$ as Gaussian:
\vspace{-0.01in}
\begin{align}
\small
& \bz\tth{t} \sim \mathcal{N}(\bs\mu(\barr\tth{t}), \bs\sigma^2(\barr\tth{t})), \\
&\bs\mu(\barr\tth{t}) = \bW_{\bs\mu}\barr\tth{t} + \bb_{\bs\mu}, \\
&\bs\sigma(\barr\tth{t}) = \mathrm{softplus}(\bW_{\bs\sigma} \barr\tth{t} + \bb_{\bs\sigma}).
\end{align}
\paragraph{Generating attentions $\&$ Training NAP} Now we generate the attention by a similar procedure to \eqref{eq:retain_attn}, but instead of feeding only $\mbf{l}\tth{1:T} = (\mbf{o}\tth{1:T}, \bh\tth{1:T})$, we feed both $\mbf{l}\tth{1:T}$ and the annotation summarization vector $\bz\tth{1:T}$ by concatenation. This allows the network to naturally reflect the information obtained from $\bz\tth{1:T}$ without having to retrain the whole attention network parameter $\bs\phi$. The original NP is meta-trained using many training examples. Likewise, NAP requires a meta-training for adapting the attention generating network $\bg_\bphi$ to take $\bz\tth{1:T}$ as an additional input~(Figure~\ref{fig:nap_concept}, (b)). We found that this adaptation requires significantly fewer training examples than the typical NP training, possibly because the network is pretrained using $\D_\text{train}$ in advance. For such adaptation training, given a set of annotated examples, we randomly subsample annotations for each training step to comprise a random task to meta-train the model. The subsampling prevents NAP from completely being over-fitted to the entire annotation set, leading to effective generalization to newly delivered annotations across rounds. We also regularize $\bz\tth{1:T}$ by positing a standard Gaussian prior distribution as in \citet{Garnelo2018np}. We train the parameters of NAP via stochastic gradient variational inference.

%, using an objective similar to the objective of the neural process.

\iffalse
NAP is trained with variational inference
Via amortized variational inference, the attention generating network $\mbf{g}_{\bs\phi}(\cdot)$ accepts new set of annotation masks (periodically delivered) as an additional input and outputs the mean and variance for the summarization vector $\bz$, reflected by the new annotation, at the constant cost of a simple forward pass.

Even with NAP, we need at least one training procedure to update $\bs\phi$ so that it can take $\bz\tth{1:T}$ as an additional input. For this training, we use two strategies to make the NAP to readily generalize to the future annotations to be given. First, at each training step, we randomly subsample the annotations to comprise random task to train the model. This prevents the model from completely being over-fitted to the entire annotation set $\{\mbf{m}_k\tth{1:T}\}_{k=1}^K$. Secondly, we regularize the summarization vector $\bz\tth{1:T}$ by positing a prior distribution. We then train the ELBO in similar fashion to the original neural process objectvie~\citep{Garnelo2018np}.
For notational simplicity, we define training points $\mbf{u}_1,...,\mbf{u}_N$, where $\mbf{u}_i = (\bx_i\tth{1:T}, \by_i)\in\D_\text{train}$\footnote{We omit the time index in this section.}.  
\fi

\newcommand{\infl}{\mathcal{I}}
\newcommand{\defeq}{\stackrel{\mathrm{def}}{=}}
\newcommand{\uval}{\bu^\text{val}}

\subsection{Cost-Effective instance and feature Reranking}
\begin{figure}[t]
% \vspace{-0.3in}
\begin{center}
\small
\begin{tabular}{c}
\includegraphics[width=7.6cm, height=3.0cm]{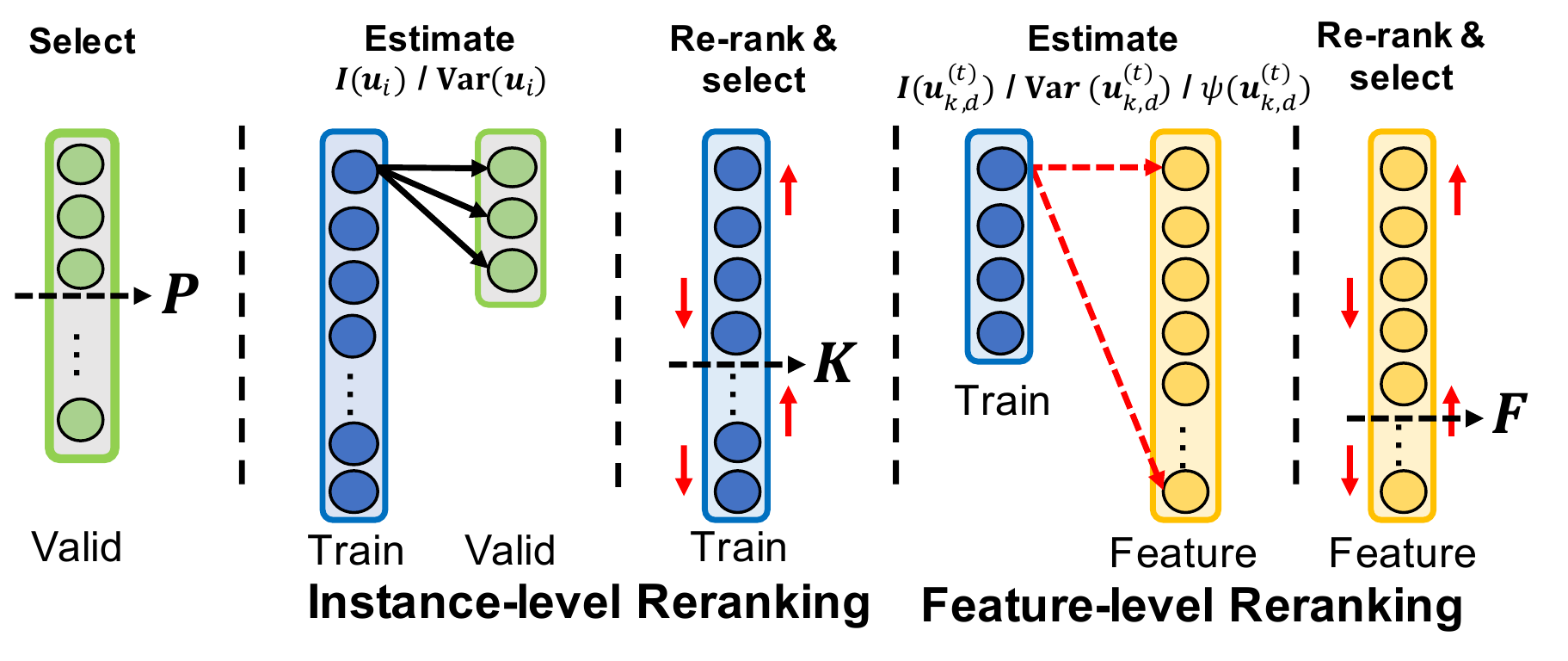} \\
\end{tabular}
\end{center}
\vspace{-0.26in}
\caption{\small Cost-Effective Re-ranking Procedure (CER).}
\label{fig:cer_figure}
\vspace{-0.24in}
\end{figure}

As we discussed earlier, letting human annotators inspect attentions for all instances and features is inefficient even for a small dataset. We may reduce this cost by randomly subsampling from
all attention values, but it may result in selecting instances or features that are already correct or have little impact to the model's prediction. Thus, we want to prioritize the attentions by their negative impact on the model's prediction, such that each feedback given by the human supervisor results in large performance improvements. In this section, we propose a general framework, depicted in Figure~\ref{fig:cer_figure},  to select important instances and features. For instance-level selection, we use \emph{the influence score and uncertainty score}. For feature-level, we use \emph{the influence score, uncertainty score, and counterfactual score}. %From now on, we abbreviate training instances $(\bx\tth{1:T}_i, \by_i)$ as $\bu_i$, ommiting the time-step indices for notational simplicity.
% \vspace{-0.06in}
\subsubsection{Instance-level reranking}
\paragraph{Influence score} 
We use the influence function~\citep{koh2017understanding} to approximate the impact of individual training points on the model prediction. The idea behind this is simple; given a validation point $\uval$, how would the validation loss change if a certain training instance $\bu$ is excluded from training procedure? Formally, let $\hat\bTheta$ be the minimizer of empirical risk for the original training set, $\frac{1}{N} \sum_{i=1}^N \loss(\bTheta, \bu_i)$, and $\hat\bTheta_{-\bu}$ be simply the one computed from empirical risk without $\bu$, $\frac{1}{N-1} \sum_{\bu_i \neq \bu} \loss(\bTheta, \bu_i)$. The effect of removing $\bu$ is then measured as $\loss(\hat\bTheta_{-\bu}, \uval) - \loss(\hat\bTheta_\bu, \uval)$. Since exactly computing this involves $N$ retraining procedures and quite expensive, \citet{koh2017understanding} propose to use the \emph{influence function} $\infl(\bu, \uval)$ to approximate it as follows:
\begin{align}
\small
&\loss(\hat\bTheta_{-\bu}, \uval) - \loss(\hat\bTheta, \uval) \approx -\frac{1}{N} \infl(\bu, \uval), \\
&\infl(\bu, \uval) \defeq - \nabla_\bTheta \loss(\uval, \hat\bTheta) H^{-1}_{\hat\bTheta} \nabla_{\bTheta}\loss(\bu, \hat\bTheta),
\end{align}
where $H_\bTheta = \frac{1}{N}\sum_{i=1}^N\nabla^2_\bTheta \loss(\hat\bTheta, \bu_i)$ is the Hessian. To summarize, the influence function $\infl(\bu, \uval)$ approximates the change in the validation loss (up to a constant) without having to retrain the model. 

During training, we are given a set of validation instances $\D_\text{valid} = \{ \uval_j \}_{j=1}^M$. Then, we first select $P$ instances that have the highest validation loss $\loss(\hat\bTheta, \uval_j)$ to 
comprise $\D'_\text{valid} = \{ \uval_p \}_{p=1}^P$. The intuition behind is that we want to select the training instances having large impact on the validation instances that are mis-predicted by the current model. In the supplementary file, we empirically show that this indeed improves the performance. Having $\D'_\text{valid}$, the influence score of a training instance $\bu_i$ is computed as $\infl(\bu_i) = \sum_{p=1}^P \infl(\bu_i, \uval_p)$.
\vspace{-0.01in}
\paragraph{Uncertainty score} 
While influence scores provide direct measures of the negative impact of an instance, it is expensive because of the Hessian computation. An alternative, and less expensive approach to measure the negative impacts is using the \emph{uncertainty}. We assume that  instances having high-predictive uncertainties are potential candidate to be corrected. This is a common approach in active learning or Bayesian optimization literature, where the points with high-uncertainties are explored. Instance-level predictive uncertainty can simply be obtained by Monte-Carlo (MC) sampling~\citep{dropout_as_bayesian}. We denote the instance-level uncertainty score as $\mathrm{Var}(\bu_i)$.
% \vspace{-0.06in}
\subsubsection{Feature-level reranking}
\paragraph{Influence score}
We can also estimate the feature-level influence score by a similar idea; if certain feature value is modified, how would the validation loss change? Let $\bu = (\bx\tth{1:T}, \by)$ be a training instance, and suppose we want to compute the influence of $\bu_{i,d}^{(t)}$, which is the $d$-th input feature for timestep $t$, $x\tth{t}_d \in \real$. Define a perturbed data point $\bu_\delta \defeq (\bx\tth{1:T} + \delta \be_{t,d}, \by)$ where $\be_{t,d}$ is an one-hot vector having $d$-th feature of $t$-th time step  as one. Let $\hat\bTheta_{\bu_\delta, -\bu}$ be the empirical risk minimizer with $\bu$ replaced by $\bu_\delta$. Then, as before, we have
\begin{align}
\small
\lefteqn{\loss(\hat\bTheta_{\bu_\delta, -\bu}, \uval) - \loss(\hat\bTheta, \uval)} \nonumber\\
&\approx -\frac{1}{N}(\infl(\bu_\delta, \uval) - \infl(\bu, \uval)).
\end{align}
Based on this approximation, we sampled $\delta$ from mean $\pm$ 2$\cdot$std of features, and computed the average influence score over multiple perturbations to rank features. As for the instance-level influence score, we add up the influence scores for all selected validation samples. We denote $\infl(\bu_{i,d}\tth{t})$ the influence score obtained by perturbing $\bu_{i,d}^{(t)}$.
% \vspace{-0.01in}
\paragraph{Uncertainty score}
NAP induces stochasticity to the attentions applied to the individual features, and this naturally leads to feature-level uncertainty scores. As for the instance-level uncertainty score, we computed variances of attentions applied for each feature by MC sampling. We denote the feature-level uncertainty score of $\bu_{i,d}^{(t)}$ as $\mathrm{Var}(\bu_{i,d}^{(t)})$.
% \vspace{-0.01in}

\begin{algorithm}[t]
\small
    \caption{Cost-Effective Re-ranking}
    \label{alg:csr}
    \textbf{\bfseries Input:} $\D_\text{train} = \{\bu_i\}_{i=1}^N$, $\D_\text{valid} = \{\uval_j\}_{j=1}^M$, $P$, $K$, $F$, $\bs\Theta\tth{s-1}$.\\
    \textbf{\bfseries Output:} $\D_\text{selection}\tth{s}=\{\bu_k\}_{k=1}^K$, $\{\balpha\tth{1:T}_k\}_{k=1}^K$.\\
    \begin{algorithmic}[1]
    \STATE Evaluate the loss for $\D_\text{valid}$.
    \STATE Sort $\{\uval_j\}_{j=1}^M$ in the descending order of $\loss(\bTheta\tth{s-1}, \uval_j)$ and select top-$P$ valid points $\D'_\text{valid}$.
    \STATE $\triangleright$ Instance-level re-ranking 
    \FOR{$i=1,...,N$}
        %\IF{Instance-rank$=$Influence}
            %\STATE Approximate $\tilde{\bI}(\bx\tth{1:T}_i)$ via \textit{Influence function}.
        \STATE Compute the influence $\infl(\bu_i)$ or uncertainty score $\mathrm{Var}(\bu_i)$.
        \STATE Select the top $K$-training points $\D_\text{selection}$ w.r.t the score.
                    %\STATE Select top $K$-training points $\D_\text{selection}\tth{s} = \{\bx_k, \by_k\}_{k=1}^{K}$ w.r.t. $\tilde{\bI}(\bx\tth{1:T}_i)$.
        %\ELSIF{Instance-rank$=$Uncertainty}
            %\STATE Approximate $\Psi(\bx\tth{1:T}_i)$ via \textit{MC dropout}.
            %\STATE Approximate $var(\bx\tth{1:T}_i)$ via \textit{MC dropout}.
            %\STATE Select top $K$-training points $\D_\text{selection}\tth{s} = \{\bx_k, \by_k\}_{k=1}^{K}$ w.r.t. $var(\bx\tth{1:T}_i)$.
        %\ENDIF
    \ENDFOR
    %\STATE Select top $K$-training points $\D_\text{selection}\tth{s} = \{\bx_k, \by_k\}_{k=1}^{K}$ w.r.t. $\Psi$ or $\tilde{\bI}$.
    \STATE $\triangleright$ Feature-level re-ranking
    \FOR{$k=1,\dots,K$}
    \FOR{$(t,d)=(1,1), \dots, (T, D)$}
    \STATE Compute influence $\infl(\bu_{k,d}\tth{t})$ or uncertainty $\mathrm{Var}(\bu_{k,d}\tth{t})$ or counterfactual $\psi(\bu_{k,d}\tth{t})$ score.
    \STATE Select top-$F$ features. 
    \ENDFOR
    \ENDFOR
    \end{algorithmic}
\end{algorithm}

\begin{figure}[t]
\vspace{-0.1in}
\begin{center}
\small
\begin{tabular}{c}
\includegraphics[width=7.0cm, height=2.9cm]{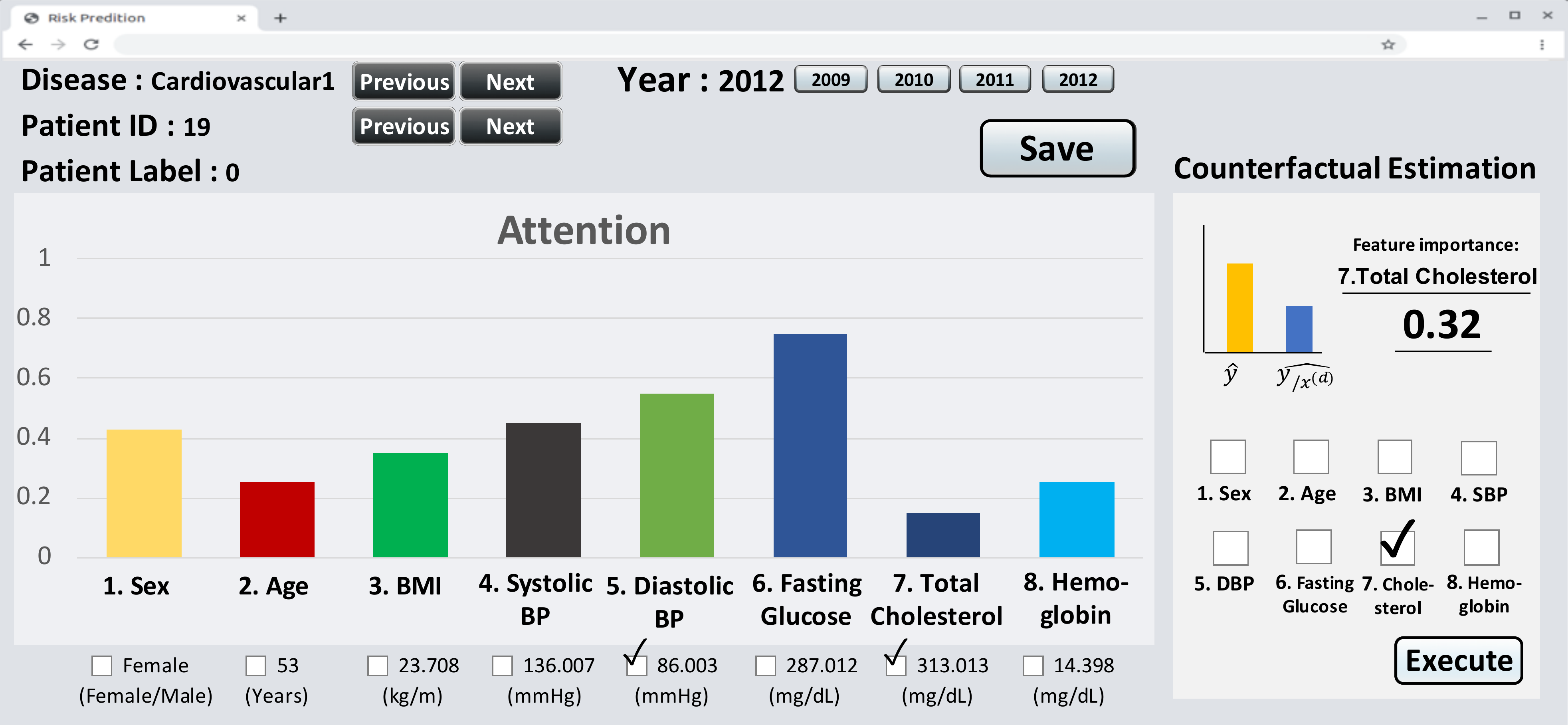} \\
\end{tabular}
\end{center}
\vspace{-0.2in}
\caption{\small Attention annotation interface (risk prediction for Cardiovascular Disease (CVD)) with counterfactual estimation tool.}
\label{fig:anno_interface}
%\vspace{-0.25in}
\end{figure}

\paragraph{Conterfactual score}
The last score, which we call as \emph{counterfactual score}, is the most direct measure of the negative impact of a feature. It answers the following question: how would the prediction change if we ignore a certain feature by manually turning off the corresponding attention value? This does not require retraining since we can simply set its attention value to zero, yet still effective because our goal is to rank the features w.r.t. their importance in attention feedback. Recall that given an attention $(\beta\tth{1:T}, \bgamma\tth{1:T})$ generated from $\bg_\bphi$, a prediction is given as 
\begin{align}
\small
\hat\by_i = \bh\bigg(\sum_{t=1}^T \beta_i\tth{t}\bgamma_i\tth{t} \odot \bv\tth{t}_i\bigg),
\end{align}
where $\bv_i\tth{1:T}$ is the linear embedding of $\bx\tth{1:T}$. The effect of perturbing $\bu_{i,d}^{(t)}$ can be then computed as follows:
\begin{align}
\small
\hat\by_{i,-(t,d)} &= \bh\bigg( \sum_{t'\neq t} \beta_i\tth{t'}\gamma_i\tth{t'} \odot \bv_i\tth{t} + \beta_i\tth{t} \bgamma_{i,-d}\tth{t} \odot \bv_i\tth{t}\bigg) \nonumber\\
\psi(\bu_{i,d}\tth{t}) &= \hat\by_i - \hat\by_{i, -(t,d)},
\vspace{-0.01in}
\end{align}
where $\bgamma_{i,-d}\tth{t}$ is the attention where $\bgamma_{i,d}^{(t)}$ = 0. We empirically found that the counterfactual score is the most effective measure for feature-level reranking (See Table~\ref{tbl:ablation_accuracy_results}).

\subsection{Human Annotation}
Finally, given a subset selected using CER whose instances and features also sorted by their negative impacts, we visualize and present the attentions to human annotators, using an online interactive user interface. We provide an example of this interface in Figure~\ref{fig:anno_interface} for the clinical risk prediction task. On the interface, the annotators set the attention mask for each feature to one of the following values: $\mbf{m}_k=\{ -1: \textit{I  don't know}, 0: \textit{Not attend}, 1: \textit{Attend}\}$. The interface visually emphasizes the features with high attentions using either a bar plot (for tabular data) or an attention map (for image data) depending on the given task. Then, the annotators examine attention weights to check whether they are incorrectly allocated, and correct them when necessary. %simultaneously checking for the corresponding input values and label. %With time-series data, annotators evaluate the delivered attentions $\alpha_\text{time}\tth{1:T}$ and $\bs\alpha_\text{variable}\tth{1:T}$ via an annotation matrix, which are basically matrices of attentions. %The same set of negative train points is delivered to multiple annotators, the accumulated sets are aggregated into one annotation matrix by averaging them: $\mbf{m}_k = \frac{1}{I}\sum_{i=1}^I \mbf m^{(i)}_k$.

\begin{table*}[t]
\begin{center}
\small
\resizebox{\textwidth}{!}{
\begin{tabular}{c|c||ccc|| c|| c}
\hline
\multicolumn{2}{c||}{\multirow{2}{*}{}}   & \multicolumn{3}{c||}{EHR}      & Fitness              & Real Estate      \\ \cline{3-5}
\multicolumn{2}{c||}{} & Heart Failure     & Cerebral Infarction & CVD     & Squat                & Forecasting     \\ \hline \hline
\multirow{3}{*}{\begin{tabular}[c]{@{}c@{}} One-time\\Training \end{tabular}} 
  & RETAIN                                            & 0.6069 $\pm$ 0.01 & 0.6394 $\pm$ 0.02   & 0.6018 $\pm$ 0.02  & 0.8425 $\pm$ 0.03  &  0.2136 $\pm$ 0.01              \\
  & Random-RETAIN                                     & 0.5952 $\pm$ 0.02 & 0.6256 $\pm$ 0.02   & 0.5885 $\pm$ 0.01  & 0.8221 $\pm$ 0.05  &  0.2140 $\pm$ 0.01              \\
  & IF-RETAIN                                         & 0.6134 $\pm$ 0.03 & 0.6422 $\pm$ 0.02   & 0.5882 $\pm$ 0.02  & 0.8363 $\pm$ 0.03  &  0.2049 $\pm$ 0.01               \\ \hline \hline
\multirow{2}{*}{\begin{tabular}[c]{@{}c@{}} Random \\ Re-ranking \end{tabular}}
  & Random-UA                                         & 0.6231 $\pm$ 0.03 & 0.6491 $\pm$ 0.01   & 0.6112 $\pm$ 0.02  & 0.8521 $\pm$ 0.02  &  0.2222 $\pm$ 0.02                \\
  & Random-NAP                                        & 0.6414 $\pm$ 0.01 & 0.6674 $\pm$ 0.02   & 0.6284 $\pm$ 0.01  & 0.8525 $\pm$ 0.01  &  0.2061 $\pm$ 0.01                \\ \hline \hline
\multirow{2}{*}{\begin{tabular} [c]{@{}c@{}} IAL\\(Cost-effective) \end{tabular}}                                                                         
  & AILA                                              & 0.6363 $\pm$ 0.03 & 0.6602 $\pm$ 0.03   & 0.6193 $\pm$ 0.02  & 0.8425 $\pm$ 0.01  &  0.2119 $\pm$ 0.01                \\
  & IAL-NAP                                           & \textbf{0.6612} $\pm$ 0.02 & \textbf{0.6892} $\pm$ 0.03   & \textbf{0.6371}  $\pm$ 0.02   & \textbf{0.8689} $\pm$ 0.01  & \textbf{0.1835} $\pm$ 0.01  \\ \hline
\end{tabular}
}
\vspace{-0.18in}
\caption{\small The binary $\&$ multi-class classification performance on the three electronic health records datasets and one fitness dataset. The reported numbers are mean-AUROC for EHR and mean-Accuracy for squat. In the real estate forecasting task, the number indicates mean-\textbf{percentage error}, meaning a lower error indicates better performance.}
\label{tbl:auc_table}
\vspace{-0.20in}
\end{center}
\end{table*}

\begin{table*}[t]
\begin{center}
\small
\resizebox{\textwidth}{!}{
\begin{tabular}{c|c||ccc|| c|| c}
\hline
\multicolumn{2}{c||}{IAL-NAP Variants}              & \multicolumn{3}{c||}{EHR}    & Fitness    & Real Estate       \\ \cline{1-5}
Instance-level & Feature-level & Heart Failure     & Cerebral Infarction & CVD     & Squat      & Forecasting       \\ \hline \hline
Influence Function & Uncertainty                & 0.6563 $\pm$ 0.01     & 0.6821 $\pm$ 0.02   & 0.6308  $\pm$ 0.02   & \textbf{0.8712 $\pm$ 0.01} &  0.1921 $\pm$ 0.01    \\
Influence Function & Influence Function         & 0.6514 $\pm$ 0.02     & 0.6825 $\pm$ 0.01   & 0.6329 $\pm$ 0.03    & 0.8632 $\pm$ 0.01   &  0.1865 $\pm$ 0.02        \\
Influence Function & Counterfactual             & 0.6592 $\pm$ 0.02     & \textbf{0.6921 $\pm$ 0.03}   & \textbf{0.6379 $\pm$ 0.02}    & 0.8682 $\pm$ 0.01  &  0.1863 $\pm$ 0.02       \\ 
Uncertainty        & Counterfactual             & \textbf{0.6612 $\pm$ 0.01}     & 0.6892 $\pm$ 0.03   & 0.6371 $\pm$ 0.02  & 0.8689 $\pm$ 0.02   &  \textbf{0.1835 $\pm$ 0.02}                 \\ \hline \hline
\end{tabular}
}
\vspace{-0.15in}
\caption{\small Results of Ablation study with proposed IAL-NAP combinations for instance- and feature-level reranking on all tasks.}
\label{tbl:ablation_accuracy_results}
\vspace{-0.1in}
\end{center}
\end{table*}
\begin{figure*}[t]
\small
    \vspace{-0.1in}
    \begin{center}
    \begin{tabular}{c c c c c}
    \hspace{-0.04in}
    \includegraphics[width=3.26cm, height=1.9cm]{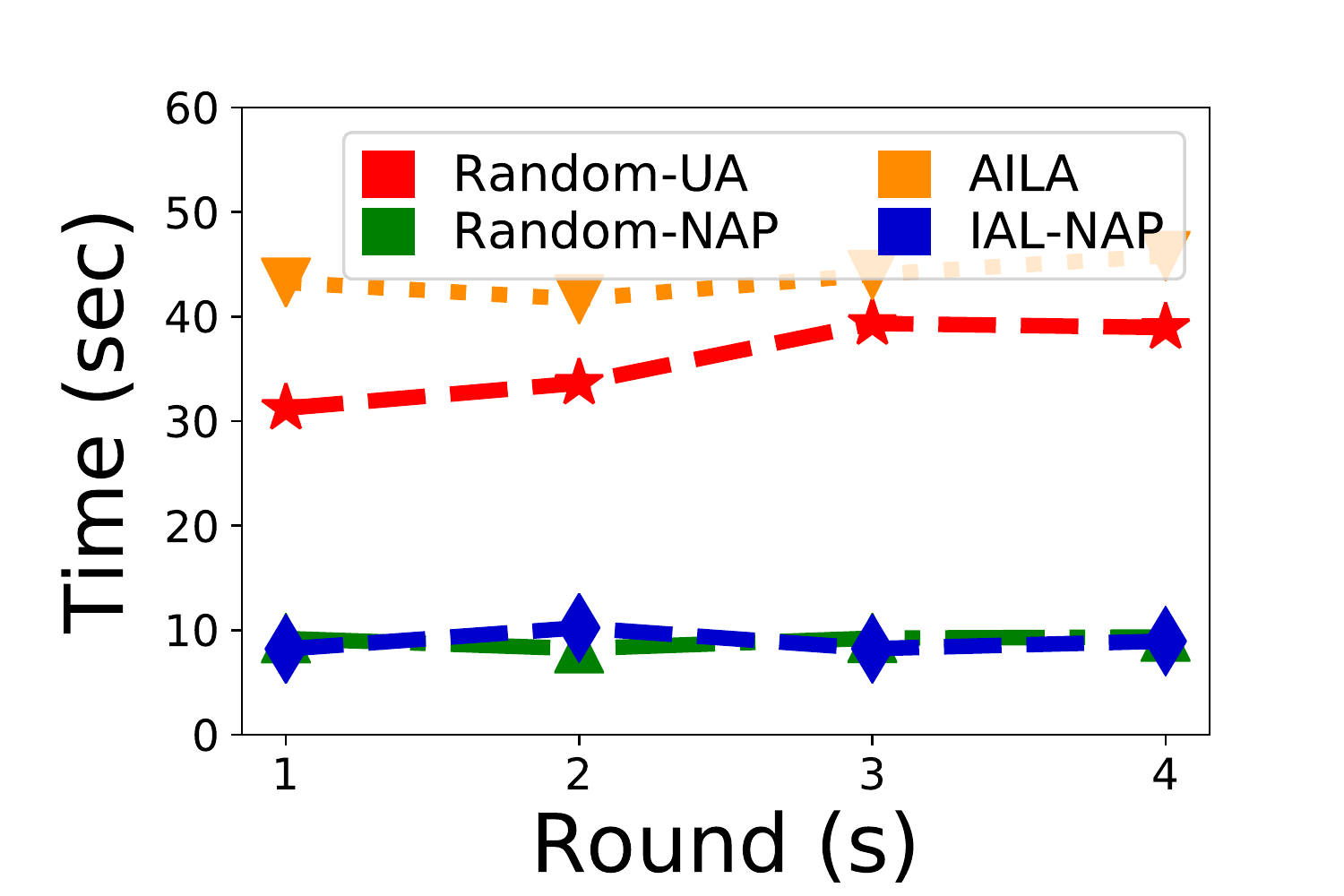}  & \hspace{-0.1in}
    \includegraphics[width=3.26cm, height=1.9cm]{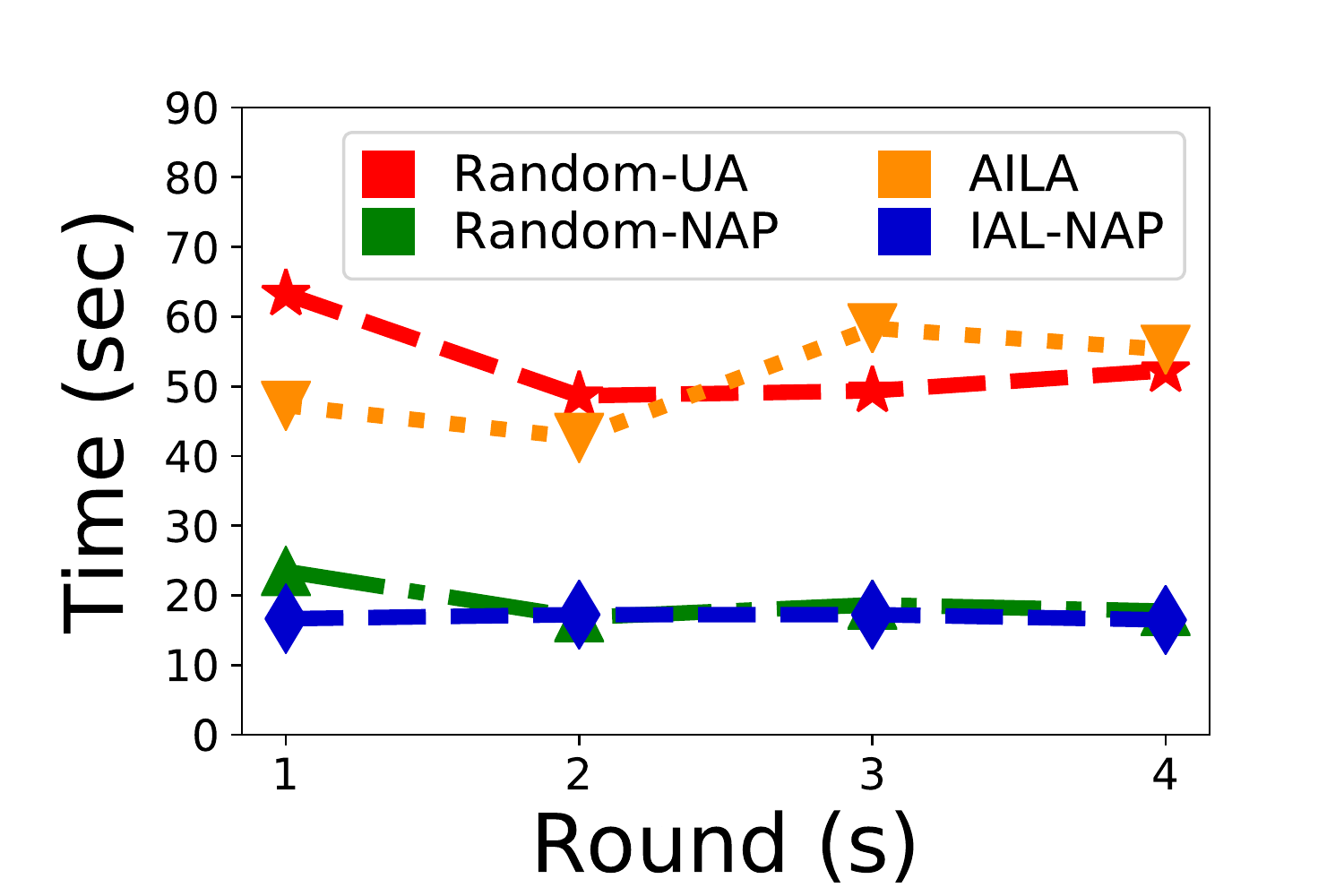}  & \hspace{-0.1in}
    \includegraphics[width=3.26cm, height=1.9cm]{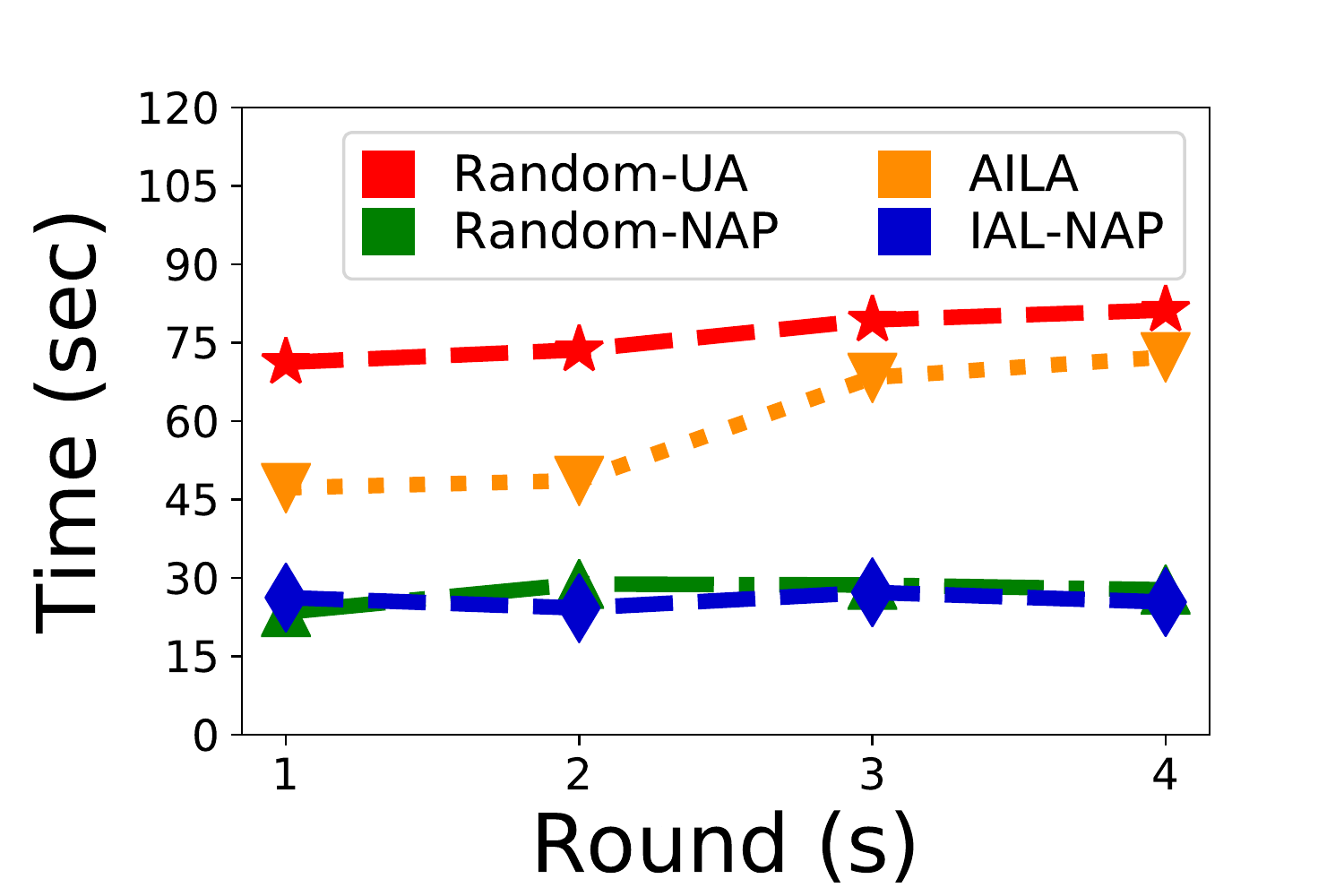}  & \hspace{-0.1in}
    \includegraphics[width=3.26cm, height=1.9cm]{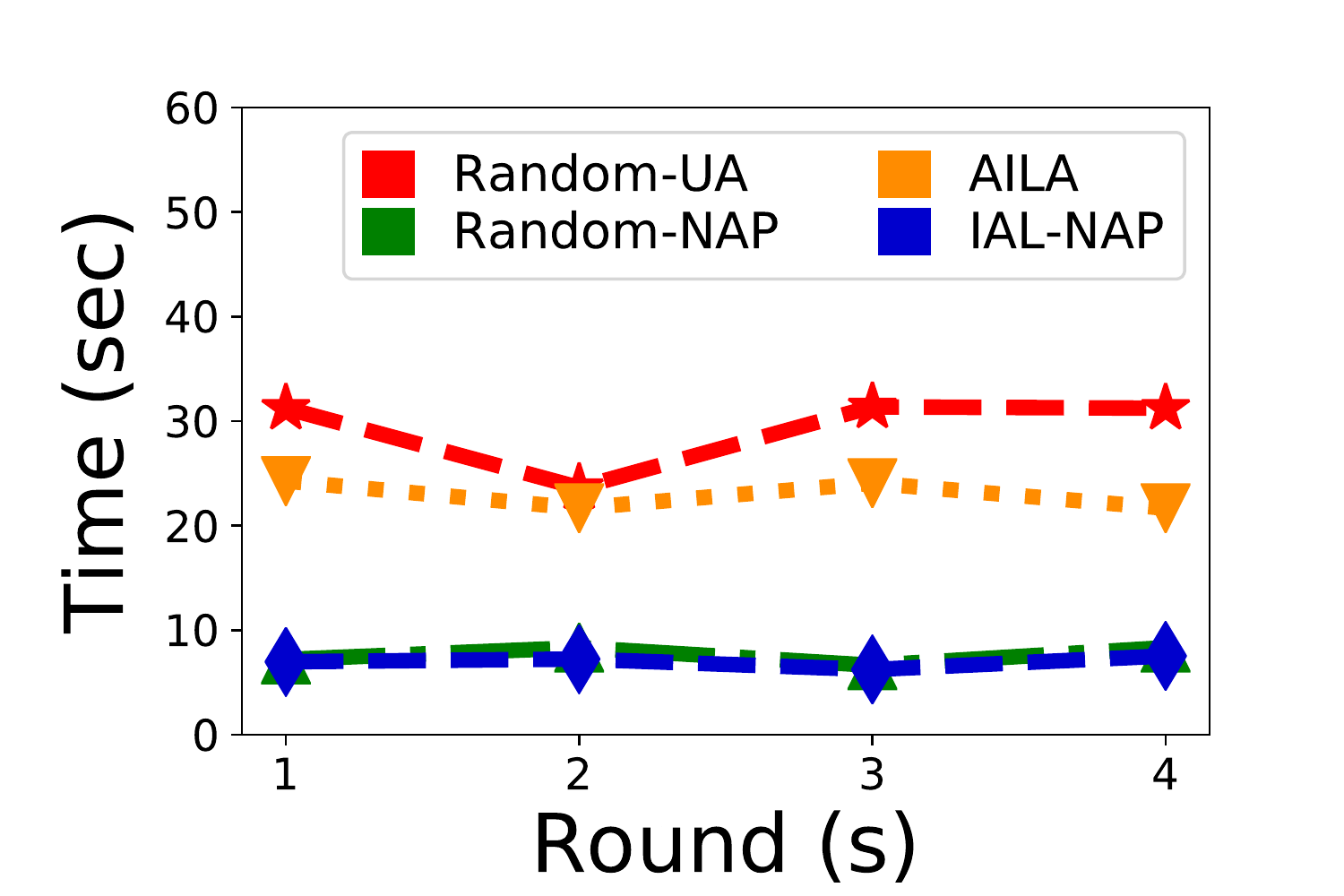}  & \hspace{-0.1in}
    \includegraphics[width=3.26cm, height=1.9cm]{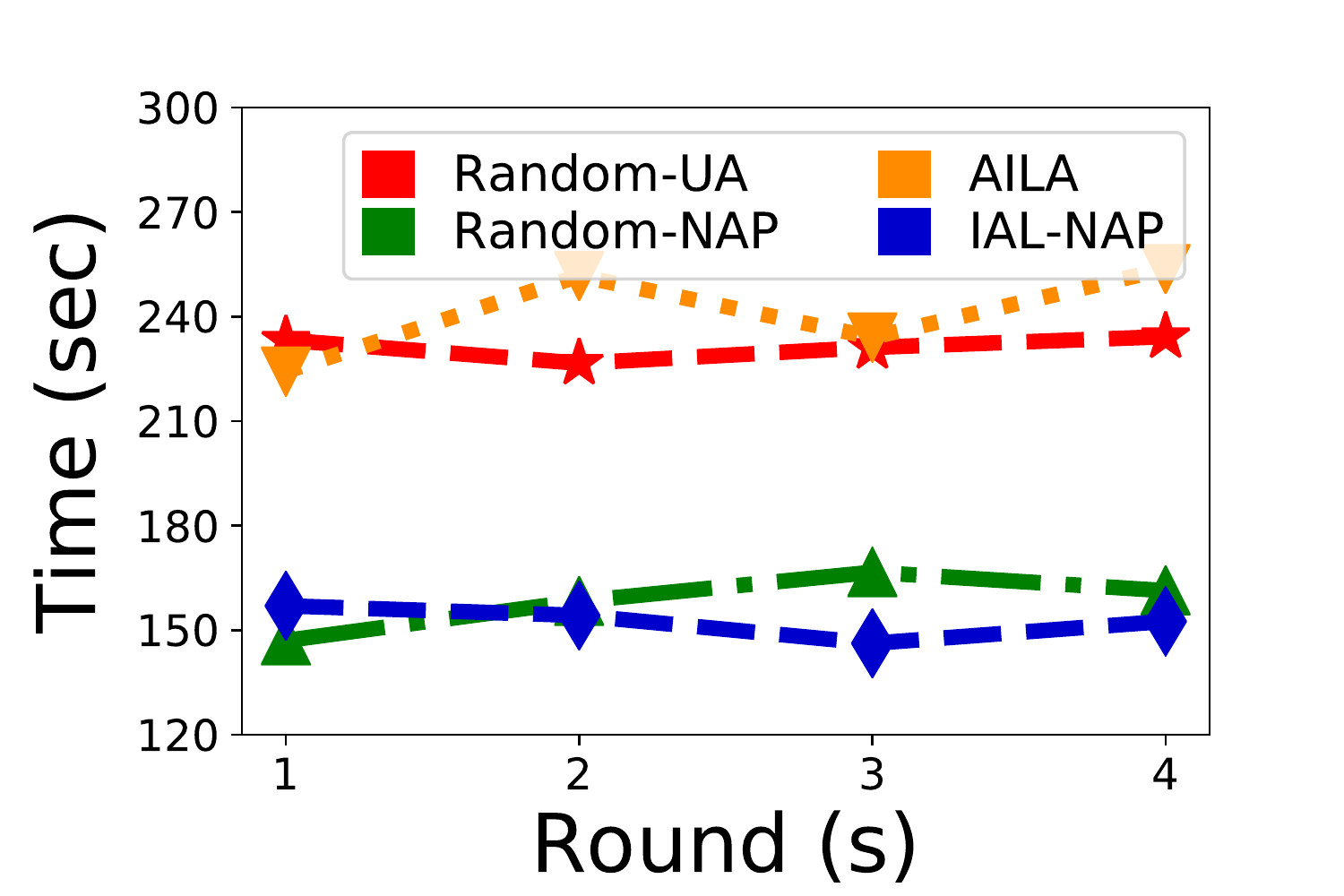}  \vspace{-0.058in} \\ 
    \hspace{-0.018in}
    \includegraphics[width=3.26cm, height=1.9cm]{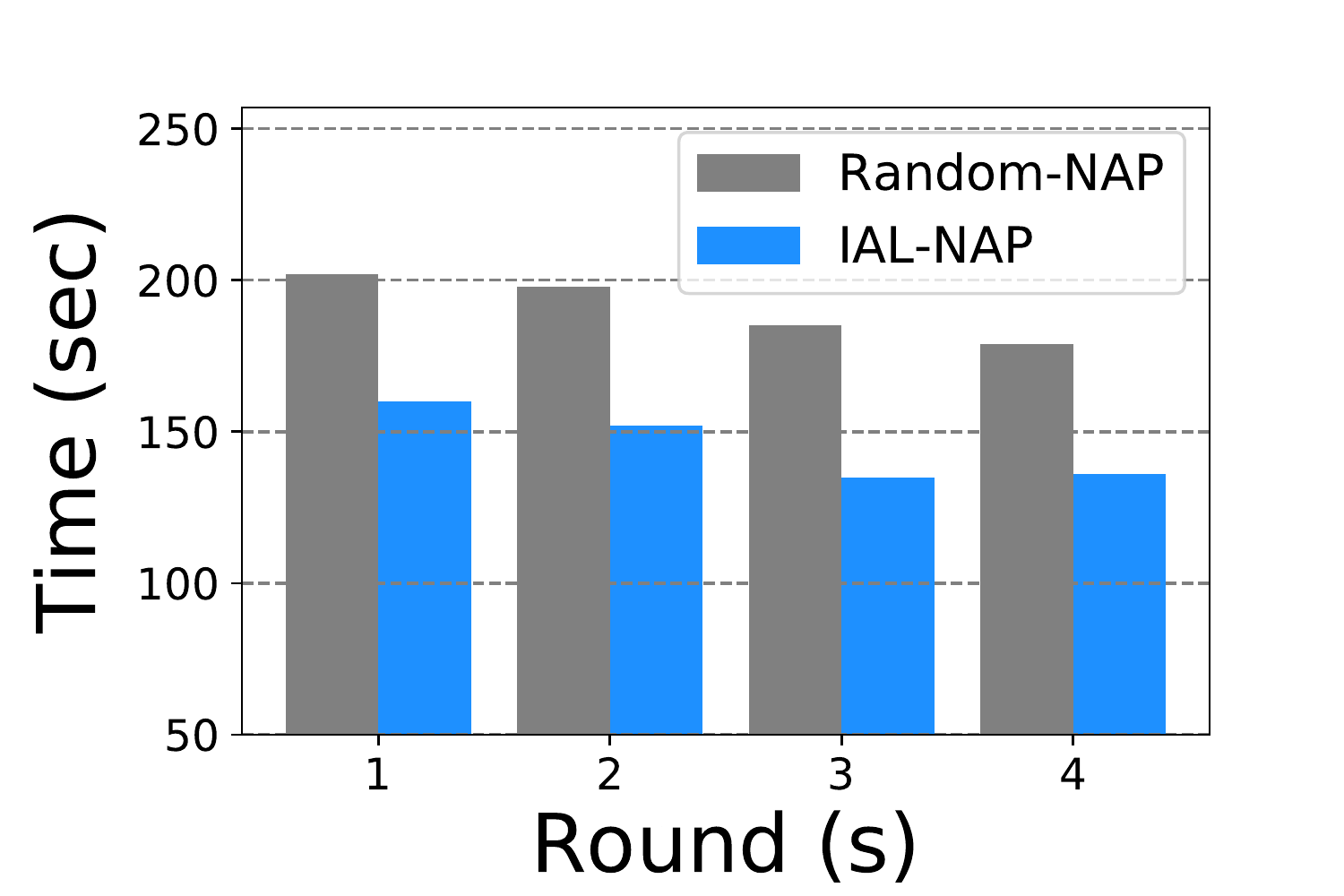}   & \hspace{-0.1in}
    \includegraphics[width=3.26cm, height=1.9cm]{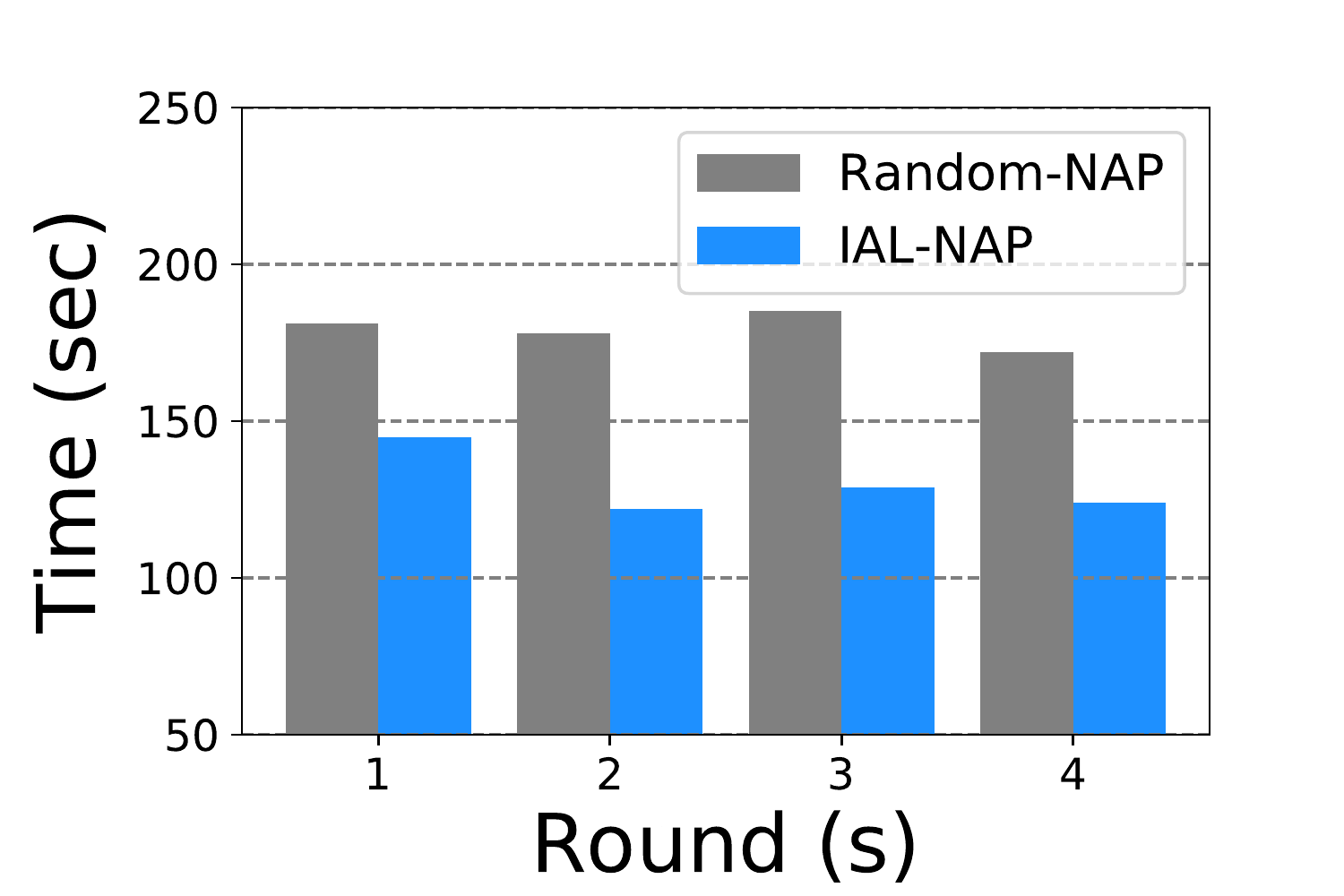}   & \hspace{-0.1in}
    \includegraphics[width=3.26cm, height=1.9cm]{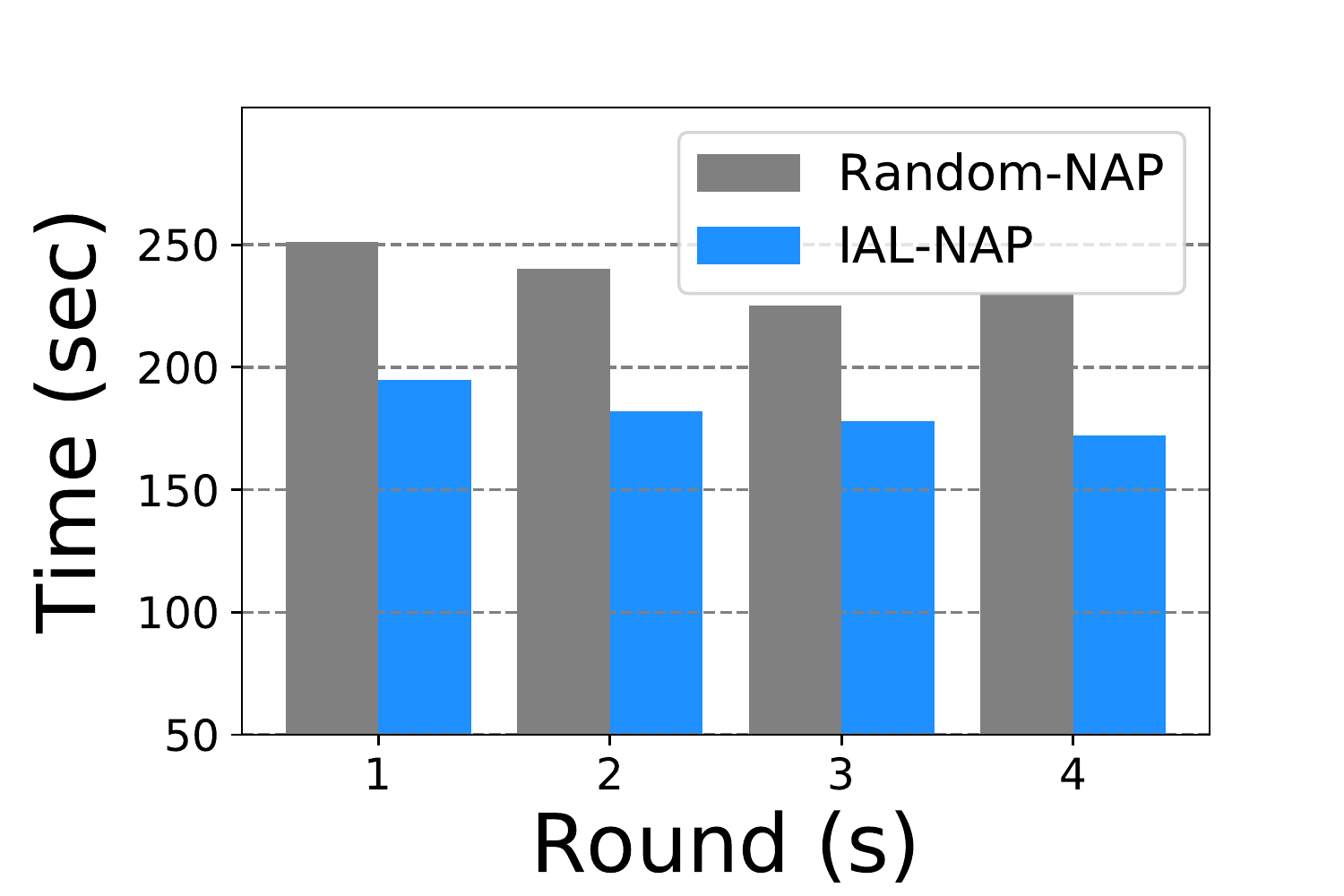}  & \hspace{-0.1in}
    \includegraphics[width=3.26cm, height=1.9cm]{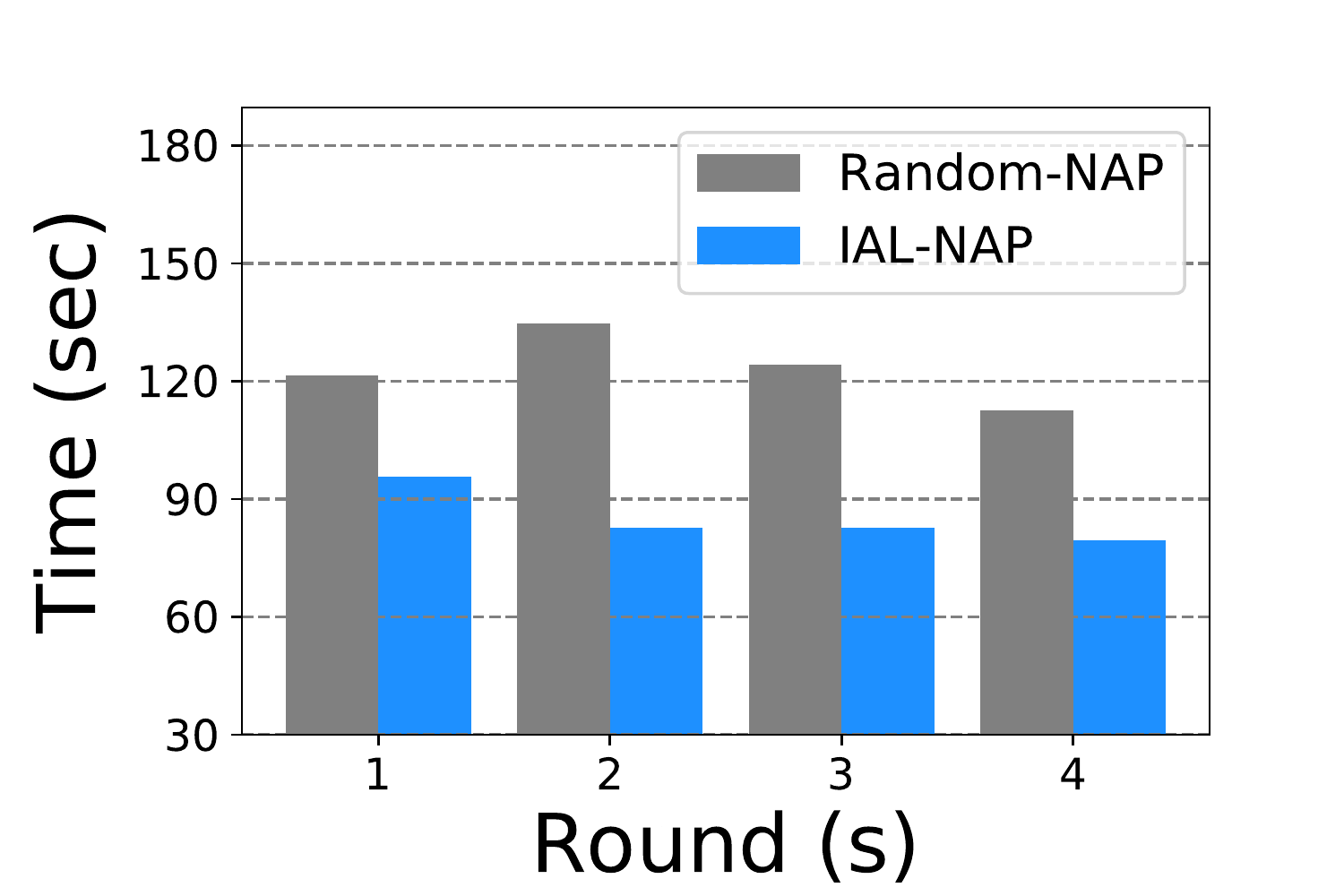}   & \hspace{-0.1in}
    \includegraphics[width=3.26cm, height=1.9cm]{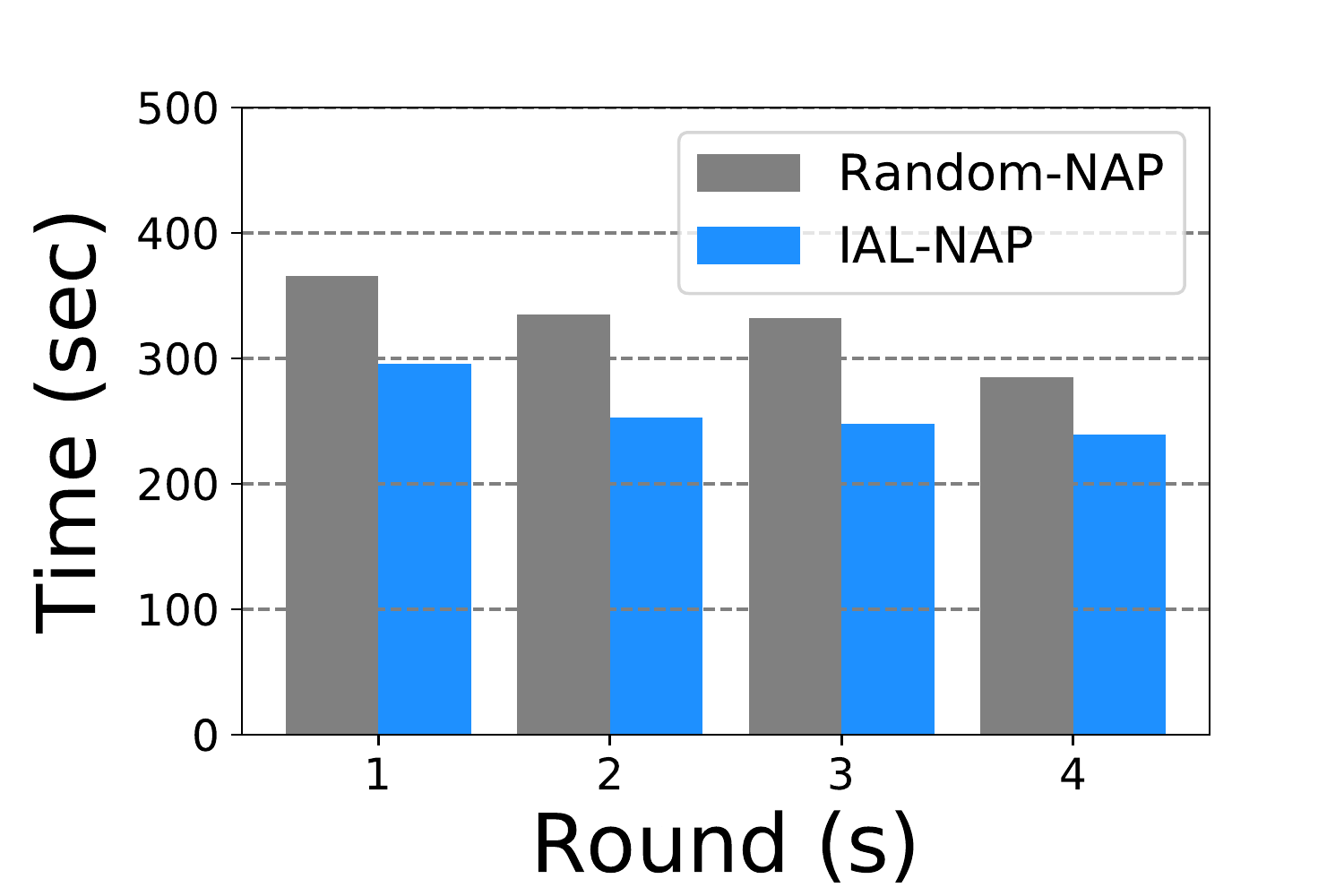}    \\
    (a) Heart Failure & (b) Cerebral Infarction & (c) CVD  & (d) Squat   & (e) Real Estate \\
    \end{tabular}
    \end{center}
    \vspace{-0.2in}
    \caption{\small \textbf{(top)} Retraining Time to retrain examples of human annotation on all task for Random-UA, AILA, Random-NAP, and IAL-NAP. \textbf{(bottom)} mean Response Time (mean-RT) of human labeling on three risk prediction task, one squat posture classification task, and one realestate forecasting task (IAL-NAP with features ranked by uncertainty \textbf{vs} Random-NAP with features ranked randomly).}
    \label{fig:time_graph}
    \vspace{-0.1in}
\end{figure*}
\vspace{-0.1in}
\begin{figure*}[ht]
\small
    \vspace{0.02in}
    \begin{minipage}{\linewidth}
      	\begin{minipage}[t]{0.62\linewidth}    	
    	\centering
    	    \resizebox{7.8cm}{!}{
    		\begin{tabular}{l | c c c c c c}
    		& Age & Smoking & SysBP & HDL & LDL\\
    		\hline
    		2009 & 31 & Yes & 139 & 54 & 97 \\
    		2010 & 32 & Yes & 134 & 55 & 97 \\
    		\textbf{Current State} & \bf33 yrs& \textbf{Yes} & \bf141 mmHg & \bf55 mg/dL & \bf102 mg/dL\\
    		\label{tbl:patient_figure}
    		\vspace{-0.15in}
    		\end{tabular}
    		}
    		\begin{tabular}{c c c}
    		\hspace{-0.2in}		
    		\includegraphics[width=3.3cm, height=2.9cm]{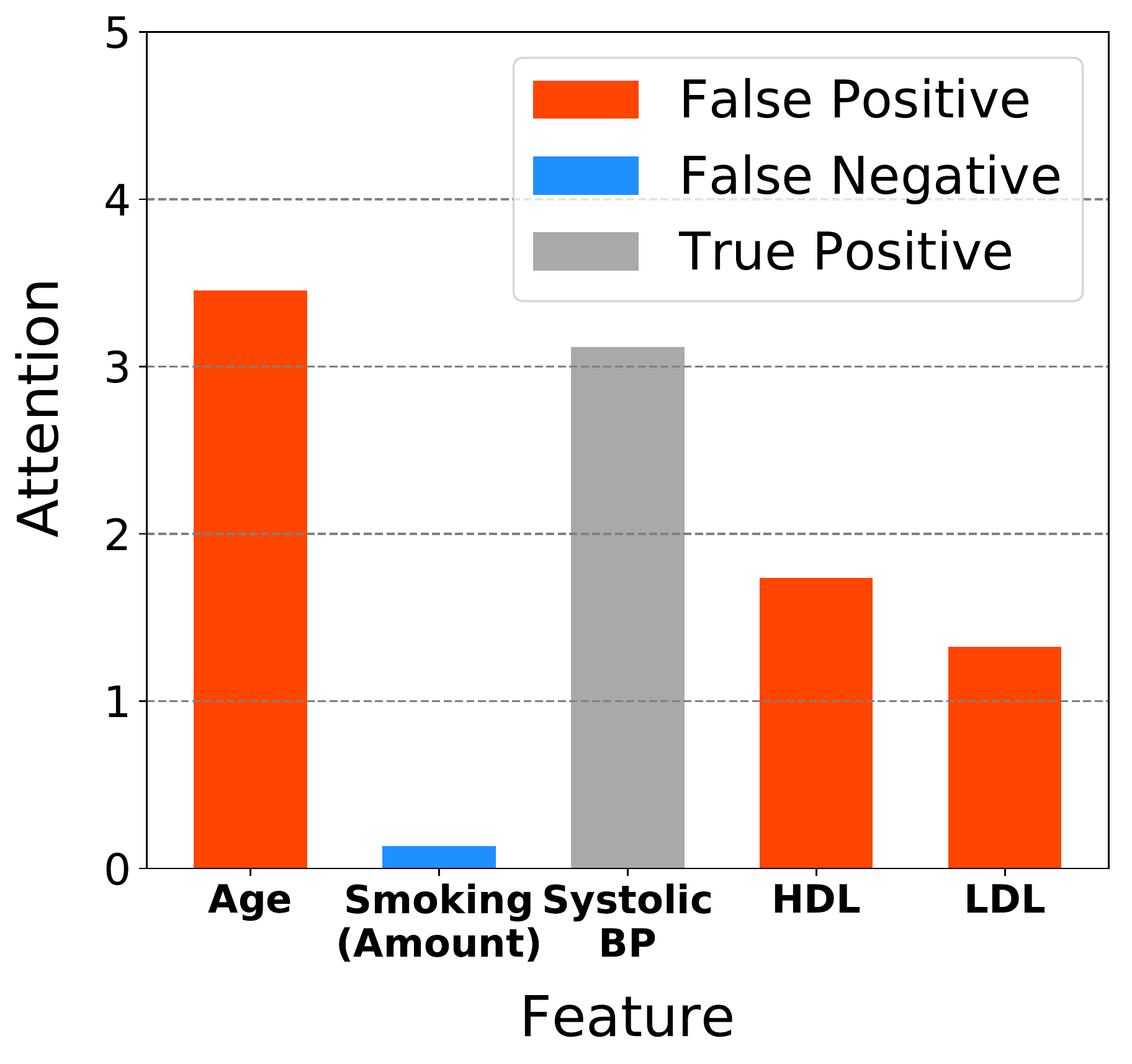}&
    		\hspace{-0.2in}
    		\includegraphics[width=3.3cm, height=2.9cm]{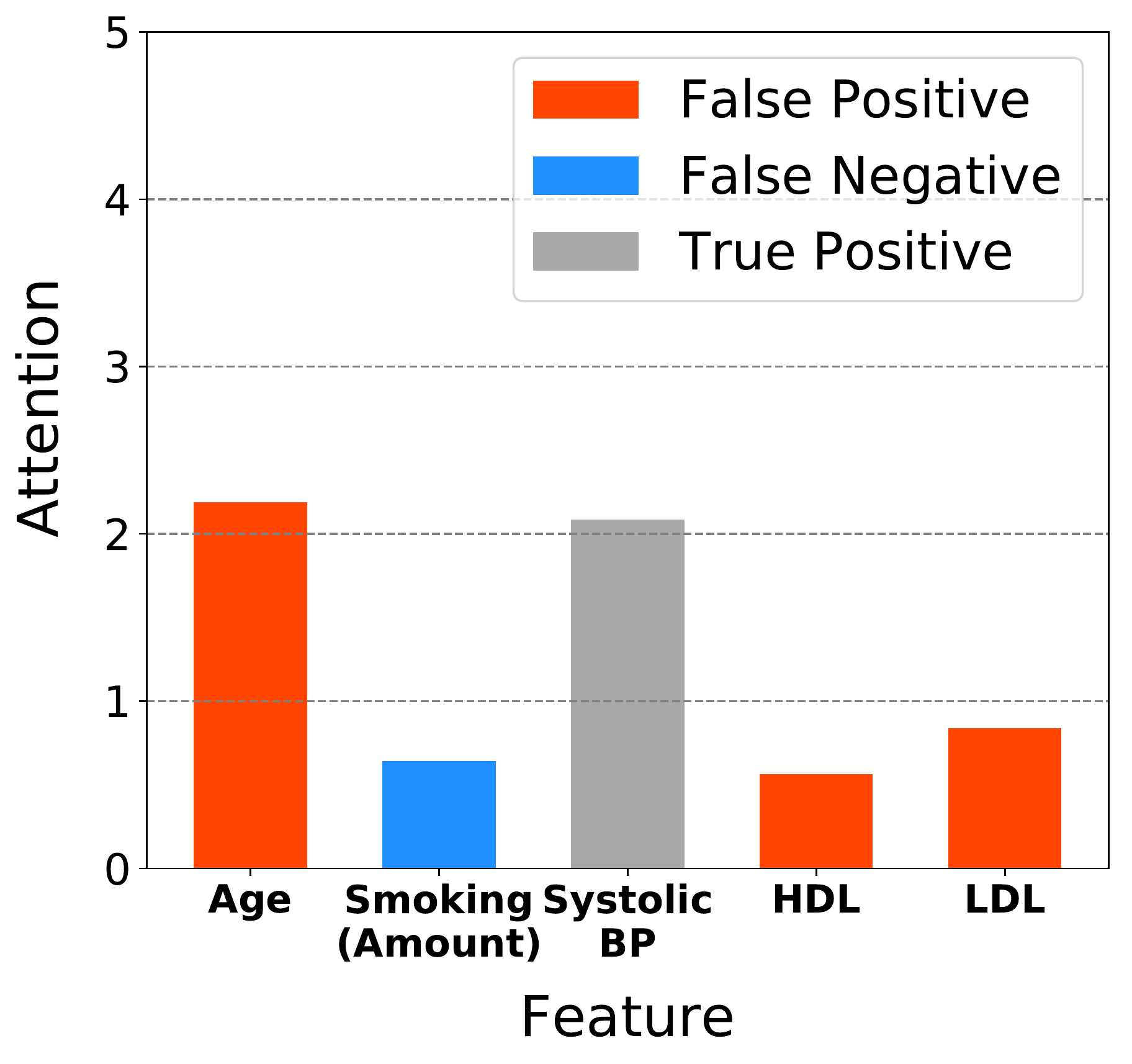}&
    		\hspace{-0.2in}
    		\includegraphics[width=3.3cm, height=2.9cm]{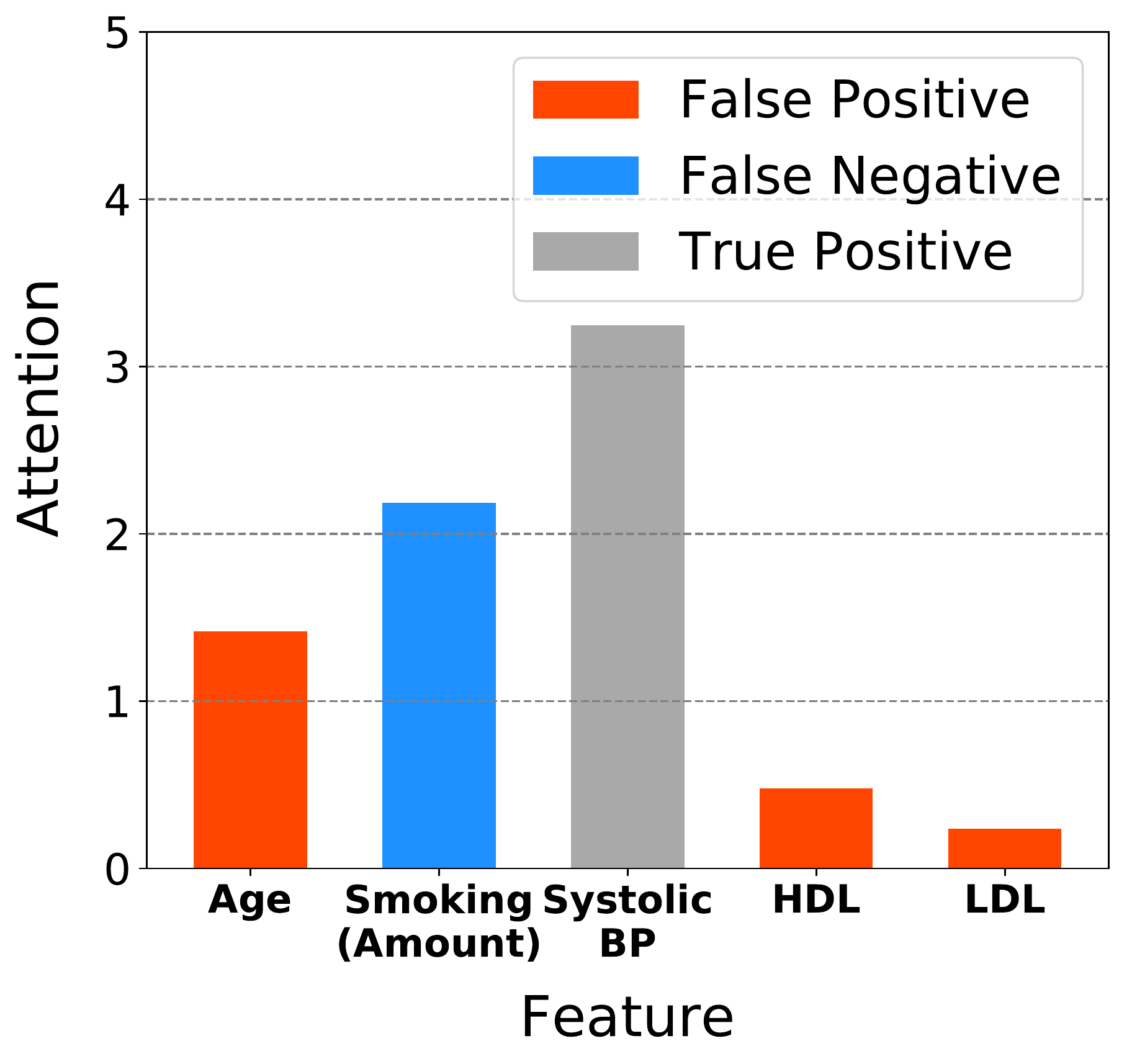} \vspace{-0.04in} \\
    		\small (a) Pretrained & (b) $s$=1 & (c) $s$=2 \\
    		\end{tabular}
    	\vspace{-0.12in}
    	\caption{\small Visualization of attention for a selected patient on Cardiovascular Disease (CVD) prediction task. Contribution indicates the extent to which each individual feature affects the onset of CVD in 1 year. \textbf{Age} - Age, \textbf{Smoking} - Whether currently smokes a cigarette, \textbf{SysBP} - Systolic blood pressure, \textbf{HDL} - High-density lipoproteins cholesterol, \textbf{LDL} - Low-density lipoprotein cholesterol. Bars correspond to attentions.}
        \label{fig:patient_qualitative_analysis}  	
      	\end{minipage}
      	\hfill
    	\begin{minipage}[t]{0.37\linewidth}
            \small
            \centering
            \begin{center}
            \vspace{-0.1in}
            \begin{tabular}{c c}
            \includegraphics[width=2.7cm, height=1.9cm]{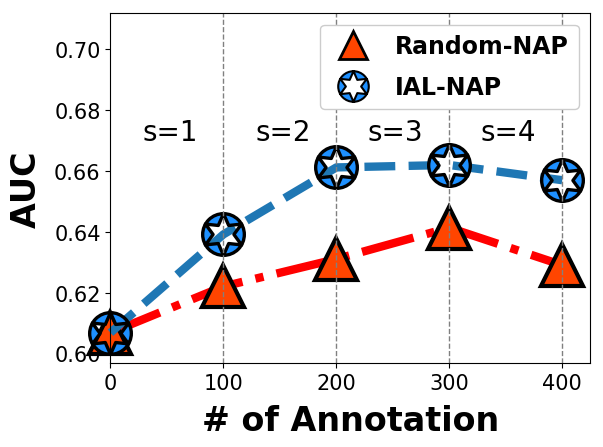}    & %\hspace{-0.01in}
            \includegraphics[width=2.7cm, height=1.9cm]{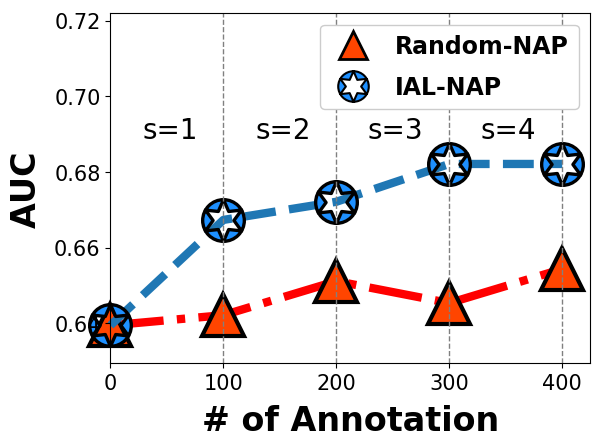} \vspace{-0.04in}\\    %\hspace{-0.01in} \\
            % \vspace{-0.01in}
            \small (a) Heart Failure & (b) Cerebral Infarction \\
            \end{tabular}
            \begin{tabular}{c c}
            \includegraphics[width=2.7cm, height=1.9cm]{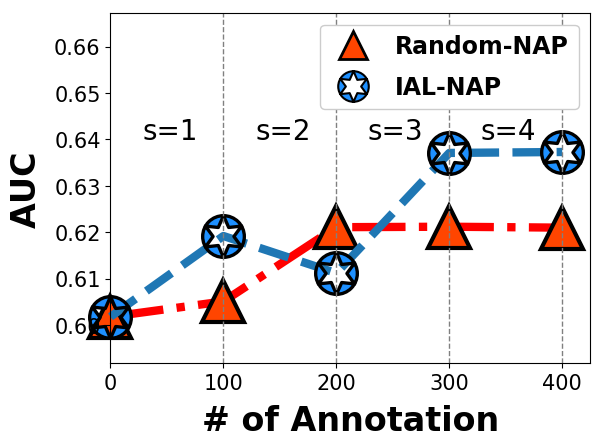}    & %\hspace{-0.01in}
            \includegraphics[width=2.7cm, height=1.9cm]{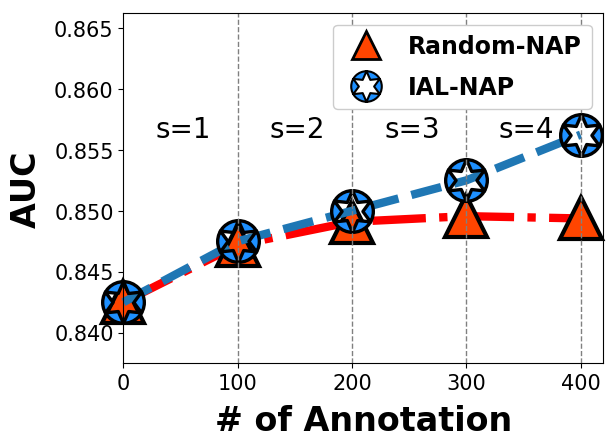}\\    %\hspace{-0.01in} \\
            \vspace{-0.08in}
            \small (c) CVD & (d) Squat \\
            \end{tabular}
    	\vspace{-0.01in}
        \caption{\small Change of accuracy with 100 annotations across four rounds ($S$) between IAL-NAP (blue) \textbf{vs} Random-NAP (red).}
        \label{fig:change_accuracy_over_stages}
            \end{center}
        \end{minipage}
    \end{minipage}
    %\vspace{-0.2in}
\end{figure*}
% \vspace{-0.05in}
\section{Experiments}
\subsection{Datasets and Baselines}
%We validate the performance and cost-effectiveness of our interactive neural attention learning framework, on five different datasets from three domains.
\paragraph{1) Medical Check-ups} These datasets are subsets of the electronic health records (EHR) database of a major hospital, which consists of medical check-ups from $2009$ to $2012$ (4 timesteps) for patients over the age of 15 in out-patient units. We extracted $245,000$ patient records from the total of $1.5$ million records, each of which contains $34$ variables including general information (e.g., sex and height), vital signs (e.g., hemoglobin level), and risk-inducing behaviors (e.g., alcohol consumption). The task is to predict the onset of the following disease in the next year: 1) \textit{Heart Failure}, 2) \textit{Cerebral Infarction}, 3) \textit{Cardiovascular Disease} (CVD). %For attention annotation procedure, we collected the responses from two clinicians of a major hospital.
%\vspace{-0.08in}
\paragraph{2) Fitness - Squat Pose Correction} This dataset contains $4,000$ video frames of human subject performing squats, where the task is to predict whether the person is performing the squat with the correct posture or with one of ten different types of incorrect postures (e.g., 0: Correct posture,  1: Exaggerated knees-forward movement, 2: Sitting on the thighs). Thus this is a multi-label classification task. We extract 14 pairs of key points from joints (e.g., \textit{left shoulder} or \textit{right ankle}) over all frames, to clearly visualize which body joints an attention generator attends to for each instance. 
%\vspace{-0.08in}
\paragraph{3) Real Estate Sales Transactions} This datasets is a subset of public rolling sales transaction database~\citep{zhu2018house} from New York City Department of Finance that is publicly available, which consists of $70,700$ house records with $27,000$ sales transaction records over 10 years from 2010 to 2019 (10 time-steps). The subset used for experiments includes $3,100$ housing transactions, each of which includes 47 variables that describes the property (e.g. number of rooms), neighborhood (e.g. minimum distance to a supermarket), and macro-economy indicators (e.g., mortgage rate). The task is to make an one-year forecast for the price of a given residential property. %Three business experts in the field of realestate $\&$ investment in New York City participated in the experiment for the annotation procedure.

\textbf{Baselines and our models} \\
\textbf{1) RETAIN:} This is the attentional recurrent neural network model (RETAIN) proposed in~\citep{retain}. \\
\textbf{2) Random-RETAIN:} RETAIN, which is newly trained from a training set without $K$ randomly selected samples.\\
\textbf{3) IF-RETAIN:} RETAIN that is newly trained from the training set without the top $K$-negative points, which are obtained using the influence function~\citep{koh2017understanding}.\\
\textbf{4) Random-UA:} This is the Uncertainty-Aware attentional network (UA)~\cite{heo2018uncertainty} which is trained using IAL with random instance and feature selection.\\ 
\textbf{5) Random-NAP:} Our IAL framework with Neural Attention Process model (NAP), which is trained using random instance and feature selection.\\
\textbf{6) Cost-effective AILA:} This is a modified version of the interactive attention learning model proposed by~\citep{aila} which retrains the attention generator by using a \textit{binary cross entropy loss function} between the attention vector $\bs\alpha_{k}$ and the attention annotation $\mbf{m}_{k}$. We train the model with CER to verify the effectiveness of the NAP.\\
\textbf{7) IAL-NAP} Our IAL framework with Neural Attention Process (NAP) and cost-effective instance and feature Reranking (CER), which uses uncertainty for instance-wise reranking and counterfactual score for feature reranking.

\textbf{Experimental setup} For all datasets, we generate train/valid/test splits with the ratio of 70$\%$:10$\%$:20$\%$. For Random-UA and AILA model, we use $\ell_2$-regularization $\|\phi^{(s)}-\phi^{(s-1)}\|_2^2$ to prevent overfitting. Please see supplementary file for more details of the datasets, network configurations, and hyperparameters. We will also publicly release the codes and all datasets used in the experiments.
%\vspace{-0.06in}
\subsection{Experimental results} We first examine the prediction performance of the baselines and our models. Table~\ref{tbl:auc_table} shows the results, where the performance is measured with \emph{Area Under the ROC curve (AUROC)} on the risk prediction tasks, \emph{accuracy} on squat posture task with multi-labels, and mean \emph{percentage error} on real estate price forecasts. Note that IF-RETAIN, which uses influence functions to remove instances with negative influence scores, performs relatively better on most tasks than other RETAIN baselines, but fails to improve on CVD and squat posture task. We observe that Random-UA, which is retrained with human attention-level supervision on randomly selected samples, performs worse than Random-NAP on all tasks. This is due to \emph{overfitting} to few supervised labels, while NAP does not suffer from overfitting.
IAL-NAP significantly outperforms Random-NAP on all tasks, which shows that the effect of attention annotation cannot have much effect on the model when the instances are \textit{randomly selected}. AILA with cost-effective reranking also performs worse than IAL-NAP, due to severe overfitting even with regularizations to prevent it. We further perform an ablation study of cost-effective reranking with different scoring measures in table~\ref{tbl:ablation_accuracy_results}. The results show that for instance-level scoring, influence and uncertainty scores work similarly, while the counterfactual score was the most effective  for feature-wise reranking. However, considering the computation cost, the combination of uncertainty-counterfactual is the most cost-effective solution since it avoids expensive computation of the Hessians.
%\vspace{-0.12in}
\paragraph{Effect of Neural Attention Process}
Line plots in Figure~\ref{fig:time_graph} \textbf{(top)} shows averaged time to retrain examples over the rounds of interactions with Random-UA, AILA, Random-NAP, and IAL-NAP on the five tasks. IAL-NAP and Random-NAP shows shorter retraining time, while Random-UA and AILA which fine-tune the attention-generating network take a longer time to retrain. This shows another benefit of our neural attention process, which is its ability to perform amortized inference. A more responsive system can also improve the quality of the interaction, in the interactive learning setting. %For retraining Random-UA and AILA, we performed early stopping to prevent overfitting and excessive retraining. 

\textbf{Effect of Cost-Effective Re-ranking}
We further measure the average response time of the annotators with and without cost-effective reranking. Figure~\ref{fig:time_graph} \textbf{(bottom)} shows that annotators spend less time with annotation if variables are prioritized by their negative impacts measured using uncertainty (blue bars) compared to presenting them in the original order (grey bars), on all tasks. Figure~\ref{fig:change_accuracy_over_stages} shows the change in model accuracy over training rounds with and without cost-effective reranking, where the negative impacts are measured by the influence score. On the risk prediction and squat posture tasks, the accuracy of IAL-NAP increases over the 4 rounds of interaction, while Random-NAP achieves only marginal increases. Especially, on the heart failure task (a), the line plot shows that IAL-NAP uses a smaller number of annotated examples (100 examples) than Random-NAP (400 examples) to improve the model with comparable accuracy (auc: 0.6414), which shows that IAL-NAP improves the model with fewer examples.
%\vspace{-0.1in}
\paragraph{Qualitative analysis}
We further analyze the contribution of each feature for a CVD patient (label=1) whose records showed significant changes in attention with the help of physicians in Figure~\ref{fig:patient_qualitative_analysis}. The table (top in Figure~\ref{fig:patient_qualitative_analysis}) shows the patient's medical records at the previous (2009, 2010) and the current time-step (2011), yearly registered records. The three graphs shows the values of the allocated attentions across three rounds. Our model, IAL-NAP failed to predict the label at  pretrained round (a), but makes a correct prediction at $s$=2 (c). We visualized five variables that have clinically meaningful changes. Across the change of attentions from (a) to (c), the physicians consider that attentions on age, HDL, and LDL in (a) are \textit{false positive} (red bars) and smoking as \textit{false negative} (blue bars), except SysBP as \textit{true positive} (grey bars). Noting that the patient's age (30) is younger than the median age (50 years-old) of female CVD patient~\citep{garcia2016cardiovascular}, initial IAL-NAP (a) allocated too much weights on age, which led to an overconfident attention model and in turn resulted in the incorrect prediction. However, our model gradually allocated less weights on age over rounds, as it started to learn \emph{what to attend to} from interactive attention learning. Note that attention on smoking highly increased at $s$=2 (c), which is also clinically guided by a physician for the reason that CVD risk increases by 25$\%$ for women who smoke cigarettes~\citep{huxley2011cigarette}. Previous incorrect attentions on HDL and LDL (a) decrease over rounds, since the HDL level (55 mg/dL) is in the normal range (40-60) and the level of LDL (102 mg/dL) is still lower than borderline high (130-159).
%\vspace{-0.04in}

\section{Conclusion} 
We proposed an interactive learning framework which iteratively learns by interacting with the human supervisors via the generated attentions. The framework utilizes a novel stochastic attention mechanism based on neural process that can correct the model's interpretation from scarce human feedback without retraining or overfitting. Further, it uses cost-effective reranking of the instances and features by their negative impacts to maximize the effect of each human-machine interaction. We validated our model on five real-world tasks from the healthcare, real estate, and fitness domains, on which our model significantly outperforms baselines with smaller retraining and human annotation cost. Qualitative analysis of our model shows that it generates more human-interpretable attentions that is crucial for its reliability on safety-critical tasks. 

\textbf{Acknowledgements}\\
This work was supported by Institute for Information $\&$ communications Technology Planning $\&$ Evaluation (IITP) grant funded by the Korea government (MSIT) (No.2017-0-01779, A machine learning and statistical inference framework for explainable artificial intelligence).


\bibliography{icml2020}
\bibliographystyle{icml2020}

% \appendix
% \section{Do \emph{not} have an appendix here}

% \textbf{\emph{Do not put content after the references.}}
% %
% Put anything that you might normally include after the references in a separate
% supplementary file.

% We recommend that you build supplementary material in a separate document.
% If you must create one PDF and cut it up, please be careful to use a tool that
% doesn't alter the margins, and that doesn't aggressively rewrite the PDF file.
% pdftk usually works fine. 

% \textbf{Please do not use Apple's preview to cut off supplementary material.} In
% previous years it has altered margins, and created headaches at the camera-ready
% stage. 
%%%%%%%%%%%%%%%%%%%%%%%%%%%%%%%%%%%%%%%%%%%%%%%%%%%%%%%%%%%%%%%%%%%%%%%%%%%%%%%
%%%%%%%%%%%%%%%%%%%%%%%%%%%%%%%%%%%%%%%%%%%%%%%%%%%%%%%%%%%%%%%%%%%%%%%%%%%%%%%

\end{document}